\pdfoutput=1
\documentclass[11pt]{article}

\usepackage{ACL2023}

\usepackage{times}
\usepackage{latexsym}
\usepackage{multirow}
\usepackage{graphicx}
\usepackage{booktabs}
\usepackage{amsmath}
\usepackage{xspace}
\usepackage{color}

\usepackage{enumitem}
\setitemize[1]{itemsep=0pt,partopsep=0pt,parsep=\parskip,topsep=5pt,leftmargin=10pt}

\usepackage[T1]{fontenc}
\usepackage[utf8]{inputenc}

\usepackage{microtype}

\usepackage{inconsolata}

\newcommand\ourkg{\textsc{PeaCoK}}
\newcommand\comfact{$\mathcal{C}{om}\mathcal{F}{act}$}
\newcommand\personachat{\textsc{Persona-Chat}}

\newcommand\atomicTT{\textsc{Atomic}$_{20}^{20}$}
\newcommand\atomic{\textsc{Atomic}}
\newcommand\comet{\textsc{Comet}}
\newcommand\psqbot{\textsc{P$^{2}$Bot}}
\newcommand\eg{\textit{e.g.}}
\newcommand\ie{\textit{i.e.}}

\newcommand\peacok{\raisebox{-8pt}{\includegraphics[width=2em]{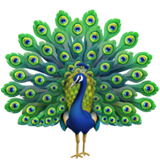}}}

\title{\peacok{} \ourkg{}: Persona Commonsense Knowledge \\ for Consistent and Engaging Narratives}

\author{\textbf{Silin Gao$^{1}$, Beatriz Borges$^{1\ast}$, Soyoung Oh$^{1\ast}$, Deniz Bayazit$^{1\ast}$,} \\
\textbf{Saya Kanno$^{2}$, Hiromi Wakaki$^{2}$, Yuki Mitsufuji$^{2}$, Antoine Bosselut$^{1\dagger}$} \\
$^1$NLP Lab, IC, EPFL, Switzerland, $^2$Sony Group Corporation, Tokyo, Japan \\
{\tt $^1$\{silin.gao,beatriz.borges,soyoung.oh,deniz.bayazit\}@epfl.ch} \\
{\tt $^2$\{saya.kanno,hiromi.wakaki,yuhki.mitsufuji\}@sony.com} \\
{\tt $^1$antoine.bosselut@epfl.ch}
}

\begin{document}
\maketitle
\renewcommand{\thefootnote}{\fnsymbol{footnote}}
\footnotetext[1]{Equal contribution.}
\footnotetext[2]{Corresponding author.}
\renewcommand{\thefootnote}{\arabic{footnote}}
\begin{abstract}
% Narrative systems (\eg{}, dialogue and storytelling systems) are often augmented with persona profiles (\eg{}, personal introductions and biographies) to learn personalized and consistent behaviours.
% However, such persona profiles are often fragmented and unorganized, which only reveal a narrow part of the integral world's commonsensical persona knowledge.

Sustaining coherent and engaging narratives requires dialogue or storytelling agents to understand how the personas of speakers or listeners ground the narrative. Specifically, these agents must infer personas of their listeners to produce statements that cater to their interests. They must also learn to maintain consistent speaker personas for themselves throughout the narrative, so that their counterparts feel involved in a realistic conversation or story.

% Sustaining coherent and engaging dialogues requires AI systems to infer an informative profile of their counterpart's personality and interests. 
% Generating coherent narratives in in stories and conversations requires AI systems to model t 
% Emulating consistent and engaging human behaviours in narratives (\eg{}, dialogues and stories) often requires natural language processing systems to understand the personas of narrators.

%In this work, we bridge this gap and propose a new large-scale persona commonsense knowledge graph, \ourkg{}, which schematizes five dimensions of persona knowledge identified in previous studies of human interactive behaviours.
However, personas are diverse and complex: they entail large quantities of rich interconnected world knowledge that is challenging to robustly represent in general narrative systems (\eg{}, a singer is good at singing, and may have attended conservatoire).
In this work, we construct a new large-scale persona commonsense knowledge graph, \ourkg{}, containing $\sim$100K human-validated persona facts. Our knowledge graph schematizes five dimensions of persona knowledge identified in previous studies of human interactive behaviours, and distils facts in this schema from both existing commonsense knowledge graphs and large-scale pretrained language models.
Our analysis indicates that \ourkg{} contains rich and precise world persona inferences that help downstream systems generate more consistent and engaging narratives.\footnote{We release our data and code to the community at \url{https://github.com/Silin159/PeaCoK}}
%Furthermore, we demonstrate that \ourkg{} can seed learned persona knowledge generators to bootstrap larger personal knowledge resources.
% Our analysis indicates that \ourkg{} contains rich and precise world persona inferences, which can seed learned persona knowledge generators.
% Experiments on dialogue modeling tasks demonstrate that \ourkg{} provides beneficial knowledge to improve persona-grounded narrative systems.
\end{abstract}

\section{Introduction}
\label{sec:intro}
Interlocutors or storytellers in narrative scenarios often exhibit varying behaviours, which are affected by their own diverse personas, but also the personas of the counterparts they are interacting with.
%In real-world conversations or storytelling scenarios, different narrators often have different behaving styles which are consistent to their diversified personas.
For example, an adventurous architect may be interested in talking about outdoor explorations with his friends who have similar hobbies, but may prefer to discuss architectural design ideas with his colleagues at work.
%To model such personalized and consistent behaviours, personal introductions, biographies and other kinds of persona profiles are incorporated into various narrative systems, promoting research in persona-grounded open-domain dialogue \citep{zhang2018personalizing,zhong2020towards,xu2022long}, story generation \cite{chandu2019my,zhang2022persona} and narrative understanding \cite{brahman2021let}.
Narrative systems must know when such behaviours should be exhibited, requiring them to learn and represent the rich personas of characters based on self-introductions, biographies and other background profiles. 
%To learn such consistent behaviours, natural language processing systems are often required to model personas of narrators based on their personal introductions, biographies and other kinds of persona background profiles, which promotes research in persona-grounded open-domain dialogue \citep{zhang2018personalizing,zhong2020towards,xu2022long}, story generation \cite{chandu2019my,zhang2022persona} and narrative understanding \cite{brahman2021let}.

\begin{figure}[t]
\centering
\includegraphics[width=1.0\columnwidth]{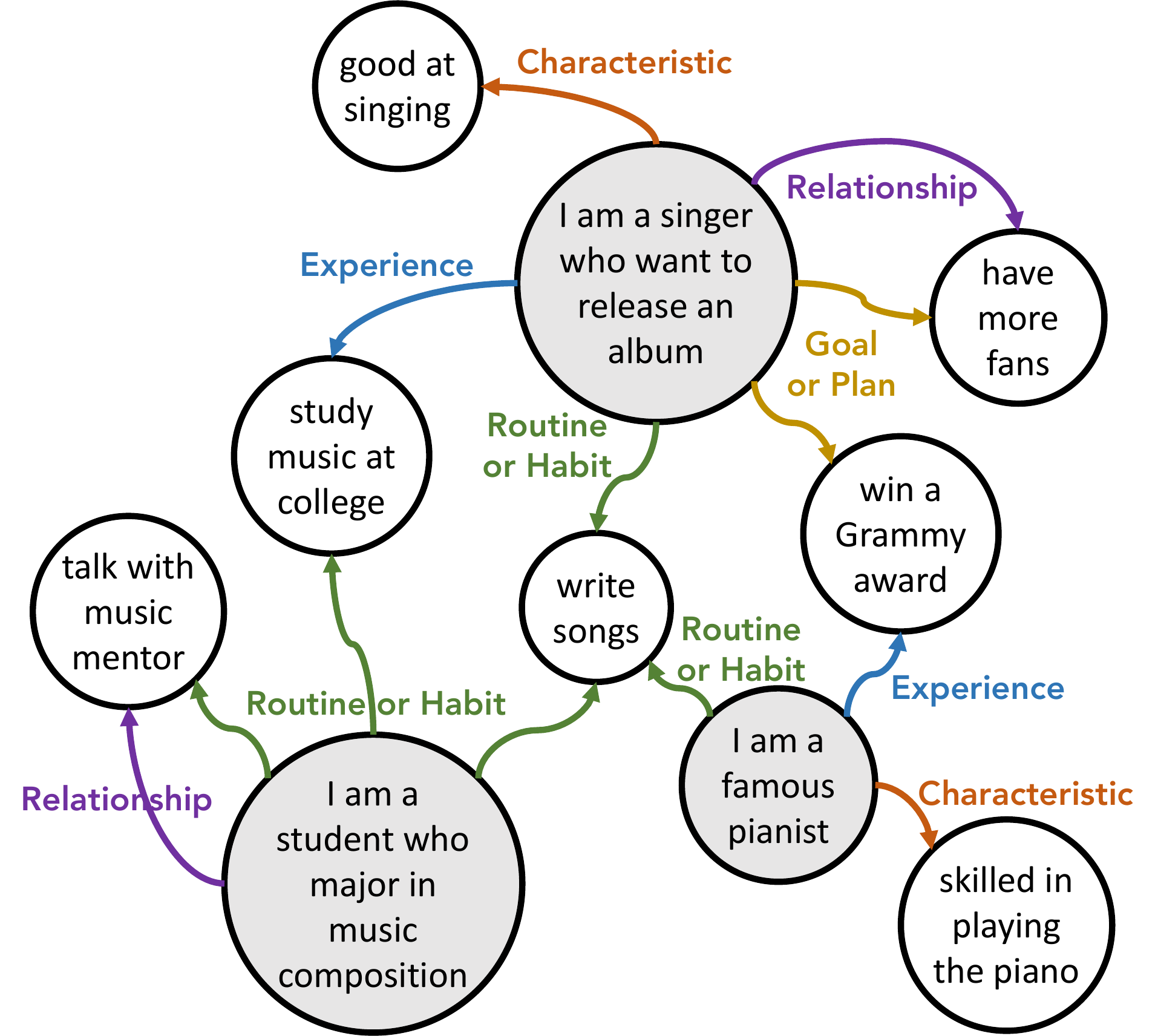}
\caption{Illustration of world persona knowledge grounded on commonsense reasoning.}
\label{fig:persona_illustrate}
\end{figure}

%However, most of the persona profiles used in current narrative systems are organized as short lists of independent persona statements.
%Each statement only presents a tiny glance of a persona, and usually has little connection with other statements, \eg{}, \textit{I released an album last month} has almost nothing to do with \textit{I love dogs very much} in the persona profile shown in Figure~\ref{fig:persona_illustrate}.
%Such fragmented profile contents make them only display a narrow view of the integral real-world personas, which involve more systematic commonsense knowledge.
This goal of modeling diverse persona attributes is at the heart of research in the areas of persona-grounded dialogue \citep{zhang2018personalizing,zhong2020towards,xu2022long}, story generation \cite{chandu2019my,zhang2022persona} and narrative understanding \cite{brahman2021let}. However, the complex nature of real-world personas, which involve rich world knowledge, and the countless ways in which they might interact, is challenging to reliably learn purely from data. 
%For instance, the person introduced by his profile in Figure~\ref{fig:persona_illustrate} is highly possible to be a professional singer, and therefore, must also be good at singing.
%But such persona knowledge is not explicitly conveyed in his fragmented and superficial profile.
For instance, as shown in Figure~\ref{fig:persona_illustrate}, a singer preparing an album may have studied music at university at one point, which would allow them to share their experience with a student majoring in composition, who may study music as a daily routine. %\antoine{}
%Such interconnected persona knowledge is usually not explicitly available in individual background profiles, and not robustly represented by language models trained on large scale data.
% Lacking access to world-level persona commonsense knowledge hinders systems from learning the systematic persona representations that underlie consistent and engaging narratives.

% Researchers have already been putting on effort to expand the persona profiles in personalized dialogue systems.
Prior work takes first steps at improving the persona knowledge representations available in narrative systems.
\citealp{mazare2018training} extract self-comments from Reddit websites to expand the scale of background persona profiles that can be used in downstream narrative settings.
However, their collected profiles are fragmented and ignore the interconnections between personas that govern interactions.
Meanwhile, \citealp{majumder2020like} use knowledge generators \citep{bosselut2019comet} to expand the persona profiles with commonsense inferences, but these commonsense expansions are limited to general social commonsense \citep{hwang2021comet}, and do not form a systematic persona-centric knowledge frame. Consequently, the lack of world-level persona commonsense knowledge resource hinders progress in learning the systematic persona representations necessary to sustain consistent and engaging narratives.

In this work, we propose a \textbf{Pe}rson\textbf{a}-grounded \textbf{Co}mmonsense \textbf{K}nowledge graph (KG), \textbf{\ourkg{}}, which represents world-level persona knowledge at scale. % grounded on commonsense reasoning.
Building off the persona concept initially proposed in human-computer interaction (\citealp{cooper1999inmates,mulder2006user,cooper2007face}) and on behaviour analysis literature for human leisure conversations \citep{dunbar1997human}, we define a \textit{persona frame} that formalizes five common aspects of persona knowledge: \textit{characteristics}, \textit{routines and habits}, \textit{goals and plans}, \textit{experiences}, and \textit{relationships}.
%Based on our designed knowledge frame, we distill persona inferences from both existing hand-crafted commonsense KGs and large-scale pretrained language models (LMs), which contributes to $\sim$100K persona-grounded facts being included in our KG.
Using this knowledge frame, we construct a large-scale graph of persona commonsense knowledge by extracting and generating persona knowledge from both existing hand-crafted commonsense KGs and large-scale pretrained language models (LMs). We validate the knowledge graph via a joint human-AI majority voting scheme that integrates large pretrained LMs into the loop of crowdsourcing, and efficiently mediates the disagreements between human annotators.

Our resulting KG, \ourkg{} contains $\sim$100K high-quality commonsense inferences (\ie{}, facts) about personas whose connectivity in the KG reveals countless opportunities to discover \textit{common ground} between personas. 
A neural extrapolation from the KG \cite{hwang2021comet} also shows that \ourkg{}'s annotated personas enable the development of effective persona inference generators.
Finally, the extended knowledge provided by \ourkg{} enables a downstream persona-grounded dialogue system to generate more consistent and engaging responses in conversations, particularly when more interconnections between the interlocutor personas are found in \ourkg{}.

% \section{Related Work}
\section{Related Work}
\label{sec:rel-work}

\noindent\textbf{Commonsense Knowledge Graphs}
Commonsense KGs such as ConceptNet \citep{liu2004conceptnet, speer2017conceptnet}, \atomic{} \citep{sap2019atomic}, \textsc{Anion} \citep{Jiang2021Anion} and \atomicTT{} \citep{hwang2021comet} are widely used in NLP applications that involve integrating implicit world knowledge, \eg{}, question answering \citep{talmor2019commonsenseqa,sap2019social,chang2020incorporating, shwartz2020unsupervised} and text generation \citep{lin2020commongen}. However, despite the importance of persona knowledge in modeling human behavior --- a crucial component for building reliable narrative systems \citep{zhang2018personalizing,chandu2019my} --- no commonsense KG explicitly focuses on representing human persona knowledge. We present \ourkg{} to open the field of developing commonsense knowledge graphs around personas. %\antoine{}

\noindent\textbf{Persona-Grounded Narratives}
%As personas of individuals contribute to capturing the background information and maintaining consistency, adapting a personality trait has become one of the crucial aspects to evaluate the success of narratives, \eg{}, dialogue system~\cite{zhang2018personalizing,zhong2020towards,xu2022long}, story generation~\cite{chandu2019my,zhang2022persona}, and narrative understanding~\cite{brahman2021let}.
%In line with this, recent works introduce persona datasets with an aim to encapsulate persona information which allows inferring information that is non-trivial without additional commonsense knowledge~\cite{mazare2018training, majumder2020like, zhang2018personalizing}.
%However, the natural and descriptive ways to build unstructured persona information (e.g., extracting persona from \textit{Reddit} comments~\cite{mazare2018training}) might result in inconsistent personas. Therefore, in this study, we introduce a systematic persona-centric knowledge framework to build a persona-grounded fact dataset.
Integrating personas to improve consistency and engagement of narratives is an important goal in dialogue \citep{song2020generating,liu2020you} and storytelling \citep{chandu2019my,zhang2022persona} systems.
One representative work that greatly contributed to the development of faithful persona emulation, \personachat{} \citep{zhang2018personalizing}, constructs a crowdsourced dialogue dataset by asking participants to perform conversations based on their assigned persona profiles --- five statements of self-introduction.
More recent work improves persona modeling in narrative systems by generating persona profiles from online resources \cite{mazare2018training}, training persona detectors \cite{gu2021detecting} and predictors \cite{zhou2021learning}, and distilling persona knowledge from commonsense inference engines \cite{majumder2020like}.
However, while these works align characters in narratives with persona profiles, they only implicitly model the areas of interaction between personas.  
In contrast, \ourkg{} explicitly represents interconnections between persona profiles, enabling persona interaction modeling in narrative systems.

\begin{figure*}[t]
\centering
\includegraphics[width=1.0\textwidth]{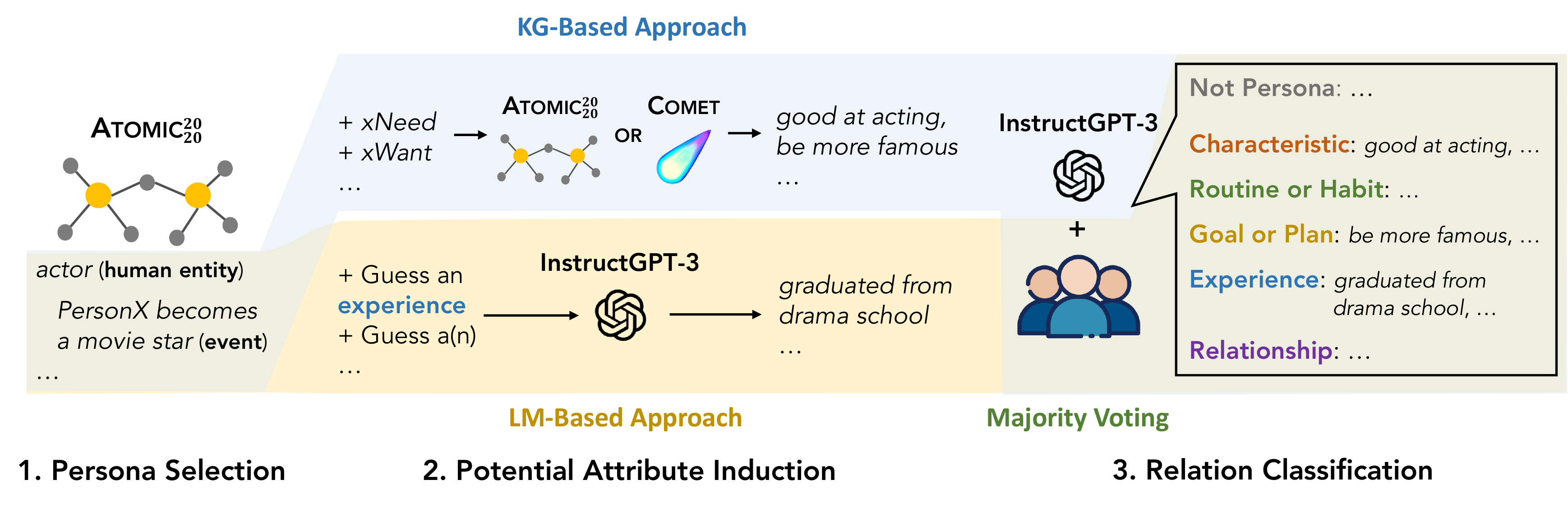}
\caption{Overview of our three-step persona-grounded commonsense knowledge graph construction.}
\label{kg_construction}
\end{figure*}

\section{\ourkg{} Knowledge Frame}
\label{sec:frame}

To construct a systematic representation of persona knowledge, we distill five common aspects of personas from classical persona definitions. % in the fields of human-computer interaction and human conversational behaviour.

% \textcolor{red}{
In the field of human-computer interaction, a persona is a fictitious example of a user group that is conceptualized to improve interactive design in areas such as marketing, communications, and service product development \citep{soegaard2012encyclopedia}. From the perspective of goal-directed design \citep{cooper1999inmates,cooper2007face}, personas encapsulate user needs and goals when interacting with a product, along with their intrinsic character traits and past experiences \citep{randolph2004usecases} that contextualize the interaction. Using these attributes of goals, traits, and experiences as the foundation of personas, we also leverage prior studies in human conversational behaviour that explore which topics of conversation are often broached in relaxed human social interactions. After conducting observational studies, \citet{dunbar1997human} categorized the topics of human conversations into bins: personal relationships (\ie{}, experiences or emotions rising from social interactions), personal experiences (\ie{}, factual events or circumstances experienced by a person), future activities (\ie{}, arrangements of meetings or events), leisure activities (\eg{}, hobbies), interests (\eg, culture, politics, religion), and work (\eg, daily routines).
% }
% \textcolor{red}{
% In the field of human-computer interaction, a persona is an archetype of a user group that is helpful to conceptualized to improve interactive design in marketing, communication and service products \citep{soegaard2012encyclopedia}. From the perspective of goal-directed design \citep{cooper1999inmates,cooper2007face}, personas consist of users' needs and goals when using a product, along with their intrinsic traits, \eg{}, emotions \citep{soegaard2012encyclopedia}, and past experiences, \eg{}, back-story \citep{randolph2004usecases}.
% Studies in HCB \citep{dunbar1997human} also found similar persona attributes driving conversational topics, which shows the main thread of interactions between interlocutors.
% The most frequent topics involved in HCB include: personal relationships (\ie{}, experiences or emotions rising from social interactions), personal experiences (\ie{}, factual events or circumstances experienced by a person), future activities (\ie{}, arrangements of meetings or events), and leisure activities (\eg{}, hobbies).
% }

To select our persona dimensions, we discard certain controversial categories from the above studies (\ie, culture, politics, and religion), as well as temporary dimensions of persona (\ie, emotion, which is well covered by prior work; \citealp{Gupta2017ASA}; \citealp{chatterjee-etal-2019-semeval}; \citealp{rashkin-etal-2019-towards}). Our final persona frame consists of five \textit{relations} for each persona, each with multiple \textit{attributes} attached to it. We describe the five relations below:

\paragraph{Characteristics} describe an intrinsic trait, \eg{}, a quality or a mental state, that the {persona} likely exhibits. For example, as shown in Figure~\ref{fig:persona_illustrate}, \textit{good at singing} describes a talent of a \textit{singer}, which is one of the singer's characteristics.
% \textcolor{red}{We derive this relation concept from the modeling of user \textit{intrinsic attributes} in HCI.}

\paragraph{Routines or Habits} describe an extrinsic behaviour that the persona does on a regular basis, \eg{}, a \textit{singer} may regularly \textit{write songs}. 
% \textcolor{red}{This relation concept is extended from the frequently involved topic \textit{hobbies} in HCB.}

\paragraph{Goals or Plans} describe an extrinsic action or outcome that the persona wants to accomplish or do in the future, \eg{}, a \textit{singer} may aim to \textit{win a Grammy award} some day.
% \textcolor{red}{This concept is mapped from the frequently involved topic \textit{future activities} in HCB and the modeling of user \textit{goals} in HCI.}

\paragraph{Experiences} describe extrinsic events or activities that the persona did in the past. For instance, a \textit{singer} may have \textit{studied music at college}. 
% \textcolor{red}{We get this relation from the frequently involved topic \textit{personal experiences} in HCB and the modeling of user \textit{historical actions} in HCI.}

\paragraph{Relationships} encode likely interactions of the persona with other people or social groups.
% , \textcolor{red}{which is derived from the frequently involved HCB topic \textit{personal relationships}.}
Note that this relation can be overlapped with other relations in \ourkg{}. For example, a \textit{singer} may want to \textit{have more fans}, which connotes a relationship between \textit{singer} and \textit{fans}, but also a future goal or plan of \textit{singer}. 

\section{\ourkg{} Construction}
We use our persona frames to construct a knowledge graph of persona commonsense where personas are treated as \textit{head} entities in the graph, frame relations constitute \textit{edge type relations}, and attributes are \textit{tails} in a (\textit{head}, \textit{relation}, \textit{tail}) structure. Then, we devise a three-step procedure to construct the frames that make up \ourkg, as shown in Figure~\ref{kg_construction}.
First, we search existing commonsense KGs to select entities that can serve as \textit{head} personas.
Then we query these KGs and prompt pretrained LMs to collect \textit{tail} attributes that are potentially associated with the personas via the five relations defined in Sec.~\ref{sec:frame}.
Finally, we use crowdsourcing with large LMs in the loop to classify whether these persona inferences are valid.

%Then we query KGs and prompt pretrained LMs based on our selected head entities, to collect fact candidates which are potentially related to the five types of personas (\ie{}, coarse-grained relations) introduced in Sec.~\ref{sec:frame}.
%Finally, we use crowdsourcing to validate the types (\ie{}, relations) of the collected persona fact candidates in a more fine-grained manner.

\subsection{Persona Selection}
\label{sec:head_selection}
We select entities that can represent \textit{head} personas using \atomicTT{} \citep{hwang2021comet}, a commonsense KG covering knowledge about physical objects, daily events, and social interactions.
We assume that entities related to personas should be about human beings, rather than other animals or non-living objects.
Therefore, we first over-sample living entities from \atomicTT{} which have animated behaviours, by extracting head entities that possess the \textit{CapableOf} relation (\ie{}, are capable of doing something), \eg{}, an \textit{actor} who is capable of performing, as shown in Figure~\ref{kg_construction}.
Then we filter out non-human beings in our extracted living entities, by removing entities that appear in the Animal Appendix of Wiktionary.\footnote{\url{https://en.wiktionary.org/wiki/Appendix:Animals}}
We also manually filter out other inappropriate entities which are too generic (\eg{}, \textit{man}) or unrealistic (\eg{}, \textit{devil}).

%Noting that \atomicTT{} has a wide coverage of social events beyond simple living entities, 
This initial procedure provides us with a diverse collection of initial coarse personas (e.g., actor, singer). To enlarge our persona set with fine-grained personas (\eg, \textit{actor who acts in movies} vs. \textit{actor who acts in plays}), %we include an additional set of event-based \textit{head} persona entities in \atomicTT{}.
%Specifically, 
we collect additional persona candidates using three types of event-based entities derived from our initial persona set: %including entities that: 
a) entities containing the initial persona in a more complex context, \eg{}, \textit{X \textbf{becomes} an actor} associates with the process of becoming an actor, rather than being an actor, b) entities that can be linked to the initial persona through the \atomicTT{} \textit{CapableOf} relation, \eg{}, \textit{X acts in play} is linked to \textit{actor}, and c) entities that are returned by Sentence-BERT retrieval \citep{reimers2019sentence} for the initial persona, \eg{}, \textit{X becomes a movie star}.
For the latter two types of derived event-based entities, we prompt InstructGPT-3 \citep{ouyang2022training} to filter out extended personas which do not entail their initial seed persona, \eg{}, \textit{X wants to be a lawyer} is not entailed by a \textit{X is a judge}, as X would already be a lawyer if they were a judge.
Finally, we extract 3.8K personas, which are converted to persona \textbf{statements} and integrated in \ourkg.% as their final representation in \ourkg{}.%, \eg{}, \textit{actor} and its derived event \textit{X becomes a movie star} is converted to \textit{I am an actor who becomes a movie star}.
\footnote{Details regarding head entity conversion and the prompt for InstructGPT-3 entity filtering are in Appendix~\ref{sec:appendix_construction}.}

\subsection{Attribute Induction}
\label{sec:tail_selection}
We derive the attribute knowledge for our collected set of head personas using both hand-crafted KGs and large language models pretrained on natural language corpora (which contain many narratives with implied persona information).

\paragraph{KG-Based Approach}
We first select 10 commonsense relations in \atomicTT{} KG which are potentially related to persona knowledge.\footnote{Appendix~\ref{sec:appendix_construction} lists our selected 10 \atomicTT{} relations and their descriptions.}
For each persona \textbf{entity} selected in Sec.~\ref{sec:head_selection}, we extract potential attributes by taking 1-hop inferences of the persona along one of our selected \atomicTT{} relations.
As \atomicTT{} may have a limited coverage of commonsense knowledge, we also use a knowledge model, \comet{} \citep{bosselut2019comet}, pretrained on \atomicTT{}, to generate potential attributes of each persona as well.
We append each selected \atomicTT{} relation to the persona entity, and feed each persona-relation pair to \comet{} to generate 5 new potential attributes.

\paragraph{LM-Based Approach}
To mine more persona knowledge implied in natural language corpora, we also prompt InstructGPT-3 to generate new persona attributes. Using each of the five relations defined in Sec.~\ref{sec:frame}, we prompt InstructGPT-3 with our persona statements and generate 5 new attributes for each relation. For example, for the \textit{Experience} relation, we instruct the model to guess distinctive activities that an individual fitting the persona might have done in the past.
We adapt InstructGPT-3 using 5 manually created in-context examples for each type of relation.\footnote{We provide our instruction and few-shot examples for InstructGPT-3 attribute generation in Appendix~\ref{sec:appendix_construction}.}

\begin{table}[t]
\centering
\resizebox{1.0\columnwidth}{!}{
\smallskip\begin{tabular}{l}
\toprule
\;\textbf{\textit{Persona}}: I am a programmer who becomes an expert\\
\textbf{Relation}: Characteristic, Self, Distinctive \\
\textbf{\textit{Attribute}}: tech savvy and highly knowledgeable in coding \\
\midrule
\;\textbf{\textit{Persona}}: I am a waiter \\
\textbf{Relation}: Routine or Habit, {Relationship}, Distinctive \\
\textbf{\textit{Attribute}}: get tips from customers \\
\midrule
\;\textbf{\textit{Persona}}: I am a runner who runs track \\
\textbf{Relation}: Goal or Plan, Self, Generic \\
\textbf{\textit{Attribute}}: get better \\
\midrule
\;\textbf{\textit{Persona}}: I am a great basketball player\\
\textbf{Relation}: Experience, Relationship, Distinctive \\
\textbf{\textit{Attribute}}: played on the varsity basketball team in high school \\
\bottomrule
\end{tabular}
}
\caption{Example persona attributes from \ourkg.}
\label{tab:example_persona}
\end{table}

\subsection{Relation Classification}
\label{sec:relation}
Once we have a large-set of initial candidate knowledge tuples to compose our persona frames, we use crowdworkers from Amazon Mechanical Turk to verify every collected relationship consisting of a \textit{head} persona, relation, and \textit{tail} attribute.
Because we observe that a fine-grained labeling schema can help workers better distinguish different relations and yield more precise annotations, we task workers with classifying fine-grained underlying features of the relations.
For each attribute, we independently ask two workers to judge whether it describes: a) an \textit{intrinsic or extrinsic} feature of the persona, b) a \textit{one-off or regular} attribute of the persona, c) a \textit{past, present or future} attribute of the persona, d) an attribute of only the persona \textit{itself}, or describing the persona's \textit{relationship} with others (\textbf{interactivity}). Finally, for each attribute in the persona frame, we ask workers whether the attribute is distinctively associated with the persona or generically associated with many potential personas (\textbf{distinctiveness}). As an example, in Table~\ref{tab:example_persona}, we see that \textit{get tips from customers} is distinctively associated as a common routine of a \textit{waiter}. Meanwhile, \textit{get better} is a generic attribute that would not be strongly associated with \textit{runner}, as many personas can have the goal of self-improvement.

\begin{figure}[t]
\centering
\includegraphics[width=1.0\columnwidth]{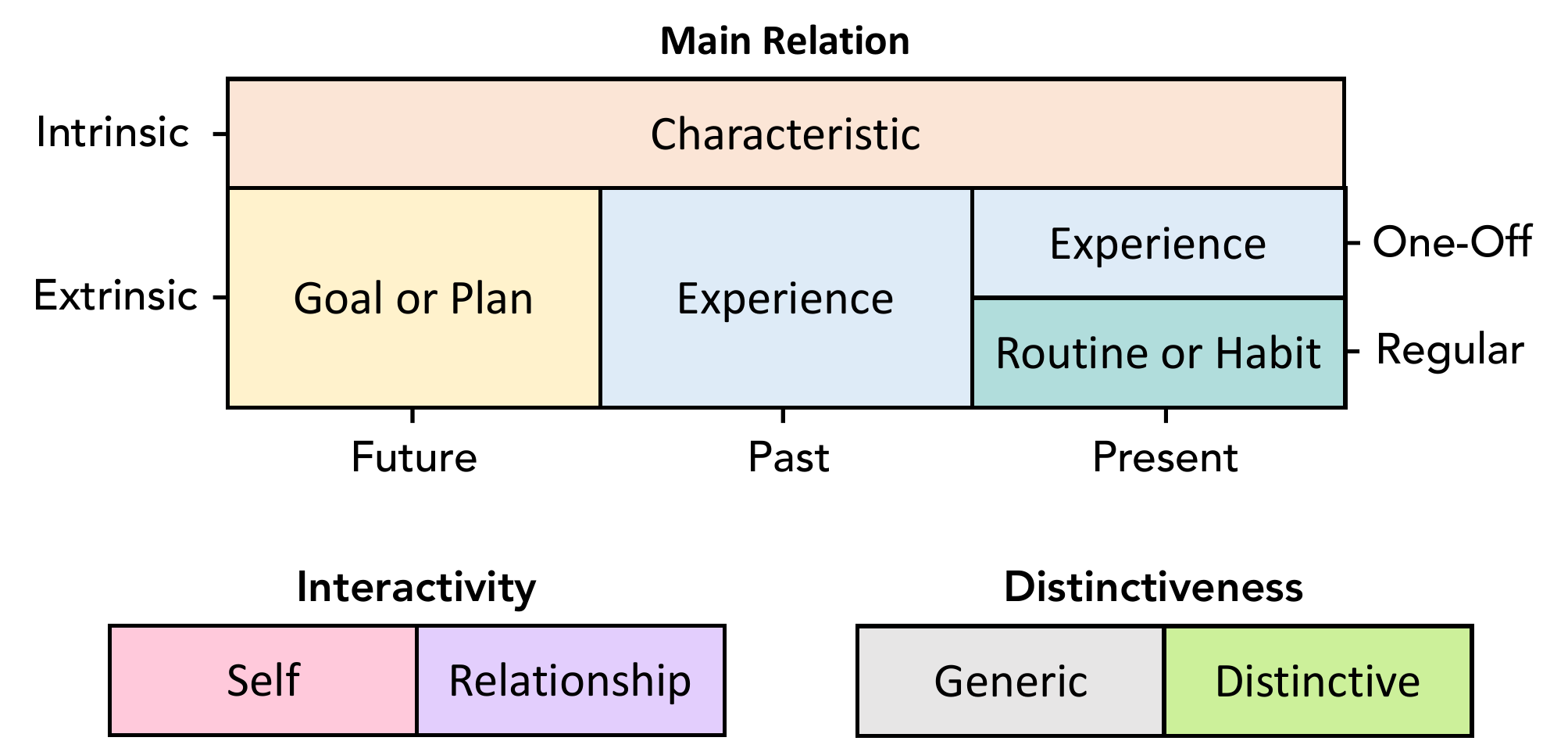}
\caption{Mapping from feature labels to relation labels.}
\label{tab:feature_relation}
\end{figure}

We follow Figure~\ref{tab:feature_relation} to map the first three dimensions of the feature labels to one of the first four relations defined in Sec.~\ref{sec:frame}, which we define as the \textbf{main} relation label of each persona-attribute pair.
The other two dimensions of feature labels, \ie{}, \textbf{interactivity} (containing the fifth relation in Sec.~\ref{sec:frame}) and \textbf{distinctiveness}, are defined as two additional relation labels.
If a worker judges that an attribute is not associated with the persona at all, we instead ask the worker to label the relation as \textbf{\textit{Not Persona}}.

% \begin{table}[t]
% \centering
% \resizebox{1.0\columnwidth}{!}{
% \smallskip\begin{tabular}{lcccc}
% \hline
% \multirow{2}*{\textbf{InstructGPT-3}} & \multicolumn{3}{c}{\textbf{Fine-Grained Labels}} \\
%                                              \cmidrule(lr){2-4}
% \multirow{1}*{\textbf{Label}} & \textbf{Main} & \textbf{Interactivity} & \textbf{Distinctiveness} \\
% \hline
% Characteristic    & Characteristic & unknown & Distinctive \\
% Routine or Habit  & Routine or Habit & unknown & Distinctive \\
% Goal or Plan      & Goal or Plan & unknown & Distinctive \\
% Experience        & Experience & unknown & Distinctive \\
% Relationship      & unknown  & Relationship & Distinctive \\
% Not Persona       & Not Persona & Not Persona & Generic \\
% \hline
% \end{tabular}
% }
% \caption{Labeling schema mapping for majority voting.}
% \label{tab:map_schema}
% \end{table}

\paragraph{Majority Voting with LM in the Loop}
To mediate the disagreements between two crowdworkers without introducing more human labour (\ie{}, a third worker), we use InstructGPT-3 and the two workers in a majority vote scheme to determine the final relation labels of some \textit{persona}-\textit{attribute} mappings.
%For each attribute collected using the KG-based approach in Sec.~\ref{sec:tail_selection}, we prompt InstructGPT-3 to produce an additional label for the relation of the attribute with respect to the persona.
For each attribute collected in Sec.~\ref{sec:tail_selection}, we prompt InstructGPT-3 to produce additional labels for the relation of the attribute with respect to the persona.
%We set the labeling classes to include the five relations defined in Sec.~\ref{sec:frame}, and also a negative class (\textbf{No Persona}) indicating that the \textit{attribute} is not a persona attribute or too generic (\eg{}, \textit{living a happy life}).\footnote{We use a simplified labeling schema instead of the fine-grained one used in crowdsourcing, because we observe that while the fine-grained annotation helped human workers frame the task, it did not help InstructGPT-3.}
We prompt InstructGPT-3 on three labeling tasks corresponding to the three dimensions of relation labeling schema shown in Figure~\ref{tab:feature_relation}.
For the \textbf{main} dimension, we set the labeling classes to include the four main relation labels, and also a negative class (\textbf{No Persona}) indicating that the \textit{attribute} is not a persona attribute or too generic (\eg{}, \textit{living a happy life}). We prompt InstructGPT-3 with 2 examples of each class for the main dimension (\ie{}, 10 manually labeled in-context examples).

For the \textbf{interactivity} and \textbf{distinctiveness} dimensions, we ask InstructGPT-3 to predict a binary label for each dimension.
%We prompt InstructGPT-3 with 2 examples of each class (\ie, 12 manually labeled in-context examples).\footnote{We include our designed instruction and few-shot examples for InstructGPT-3 relation labeling in Appendix~\ref{sec:appendix_construction}.}
% For each persona attribute collected by the LM-based approach, InstructGPT-3's vote is the relation type considered during its generation.
For these predictions, we provide InstructGPT-3 with 4 examples of each class (\ie{}, 8 manually labeled in-context examples for each dimension).\footnote{We include our designed instruction and few-shot examples for InstructGPT-3 relation labeling in Appendix~\ref{sec:appendix_construction}.}

%We follow Table~\ref{tab:map_schema} to map InstructGPT-3's labels into the fine-grained schema used in crowdsourcing.
%Then on each dimension of the fine-grained labeling schema, we determine the final label as the majority label given by InstructGPT-3 and the two workers.
For each dimension of the relation labeling schema shown in Figure~\ref{tab:feature_relation}, we determine the final label as the majority label given by InstructGPT-3 and the two workers. 
We set the final label as \textbf{\textit{Controversial}} if no unique majority label is found, \eg{}, InstructGPT-3 and two workers all give different labels.
Finally, each \textit{persona}-\textit{attribute} pair forms a persona fact triple with its annotated relation labels in \ourkg{}.
Table~\ref{tab:example_persona} shows some examples of \ourkg{} facts.\footnote{We list more \ourkg{} persona facts in Appendix~\ref{sec:appendix_examples}.}
%To get a pretrained LM's view on distinguishing different types of persona knowledge (besides human perspective), we prompt InstructGPT-3 to label (\ie{}, classify) the persona-centric relations between each \textit{head} and its potential \textit{tail}.
%\footnote{We use different labeling schemas for InstructGPT-3 and crowdworkers to fit their different reasoning styles. For InstructGPT-3, we find that simplified classification or generation procedure can promote model performance, so we set different relation types to be independent of each other. While for crowdworkers, we observe that the fine-grained labeling schema can help workers better distinguish the concepts of different relations and give more precise annotations.}

\begin{table}[t]
\centering
\resizebox{1.0\columnwidth}{!}{
\smallskip\begin{tabular}
{lccc}
\toprule
\multirow{2}{*}{\textbf{Dimension}} & \multirow{2}{*}{\textbf{Type}} & \multicolumn{2}{c}{\textbf{Approach}} \\
\cmidrule(lr){3-4} 
& & \textbf{KG-Based} &\textbf{LM-Based}\\
\toprule
\multirow{8}{*}{Main} &  \multirow{2}{*}{Characteristic}  &  9133 & 13033 \\
&  & 22.5\%  & 21.2\% \\
\cmidrule(lr){2-4}
& \multirow{2}{*}{Routine/Habit} & 22991 & 24461 \\
& & 56.5\% & 39.8\% \\
\cmidrule(lr){2-4}
& \multirow{2}{*}{Goal/Plan} & 3368 & 11447 \\
& & 8.3\% & 18.6\% \\
\cmidrule(lr){2-4}
& \multirow{2}{*}{Experience} & 5171 & 12493 \\
& & 12.7\% & 20.3\% \\
\midrule
\multirow{4}*{Interactivity} & \multirow{2}{*}{Relationship} & 6990  & 17503 \\
& & 17.2\% & 28.5\% \\
\cmidrule(lr){2-4}
& \multirow{2}{*}{Self} & 33673 & 43931 \\
& & 82.8\% & 71.5\% \\
% \cmidrule(lr){2-4}
% & \multirow{2}{*}{Controversial} & 7352 & 10332\\
% & & 18.6\% & 17.4\% \\
\midrule
\multirow{4}*{Distinctiveness}  & \multirow{2}{*}{Distinctive}  &  26413  & 56741 \\
& & 65.0\% & 92.4\% \\
\cmidrule(lr){2-4}
& \multirow{2}{*}{Generic} & 14250 & 4693 \\
& & 35.0\% & 7.6\%\\
\midrule
\textbf{Total} & & 40663 & 61434 \\
\bottomrule
\end{tabular}
}
\caption{Statistics of persona relations in \ourkg{}. %We exclude persona facts whose labels in the main dimension are \textit{Controversial} (\textit{N}=8083+2830) or \textit{Not Persona} (\textit{N}=2189+138), 
%We include facts whose labels in the interactivity dimension are \textit{Controversial}, since they retain meaningful main relation labels.
}
\label{tab:final_labels}
\end{table}

\begin{table*}[t]
\centering
\resizebox{1.0\textwidth}{!}{
\smallskip\begin{tabular}
{llccccc}
\toprule
\multirow{2}*{\textbf{Dimension}} & \multirow{2}*{\textbf{Label}} & \multicolumn{4}{c}{\textbf{Workers Disagree}} & \multirow{2}*{\textbf{Workers Agree}} \\
\cmidrule(lr){3-6} 
& & \textbf{GPT3 \& W1} & \textbf{GPT3 \& W2} & \textbf{Controversial} & \textbf{Total} & \\
\toprule
\multirow{5}*{Main} & Characteristic &  3770 (9.2\%) &  4194 (10.2\%) &  \multirow{5}{*}{10913 (26.5\%)}  & \multirow{5}{*}{41161} & \multirow{5}{*}{71849}\\
\cmidrule(lr){2-4} 
% \midrule
& Routine or Habit &  4506 (10.9\%)  &  3265 (7.9\%) & &\\
\cmidrule(lr){2-4}
& Goal or Plan  & 4786 (11.6\%) &  3458 (8.4\%) & &\\
\cmidrule(lr){2-4}
& Experience  & 3457 (8.4\%) &  2812 (6.8\%) & &\\
\midrule
\multirow{2}*{Interactivity} & Relationship & 4933 (23.6\%) &  5382 (25.7\%) & \multirow{2}{*}{-} & \multirow{2}{*}{20940} & \multirow{2}{*}{81157} \\
\cmidrule(lr){2-4}
                             & Self         & 4657 (22.2\%) &  5968 (28.5\%) &  &  &  \\
\midrule
\multirow{2}*{Distinctiveness} & Distinctive & 16790 (49.2\%) &  8011 (38.3\%) & \multirow{2}{*}{-} & \multirow{2}{*}{34135} & \multirow{2}{*}{67962} \\
\cmidrule(lr){2-4}
                               & Generic     &  2475 (7.3\%)  &  6859 (32.8\%) &  &  &  \\
\bottomrule
\end{tabular}
}
\caption{Statistics of labeling disagreements. \textbf{GPT3 \& W1}: InstructGPT-3 and the first worker agree on the final labels, \textbf{GPT3 \& W2}: InstructGPT-3 and the second worker agree on the final label, \textbf{Controversial}: No agreement between InstructGPT-3 and either of the two workers, resulting in the final label being \textit{Controversial}. Percentage values in parentheses are computed among cases where there is disagreement between the two workers.}
\label{tab:final_disagree}
\end{table*}

\section{\ourkg{} Analysis}
\label{sec:data}
Our statistics of the final \ourkg{} relations are shown in Table~\ref{tab:final_labels}, where we construct 102,097 facts with valid persona knowledge inferences.
We stratify \ourkg{} statistics based on the two persona collection approaches (KG-based and LM-based) described in Sec.~\ref{sec:tail_selection}.
We find that the KG-based distillation (which extracts information initially annotated by human workers) results in more imbalanced persona knowledge. A large proportion ($\sim$57\%) of \textit{Routine or Habit} relations dominate the extracted persona relations, and there are fewer \textit{Relationship} and \textit{Distinctive} facts, as well.
This indicates that hand-crafted social commonsense KGs contain a narrower view of real-world persona knowledge, highlighting the importance of also distilling a balanced set of persona knowledge from large pretrained LMs.
However, the repurposed knowledge from the KG was initially written by humans, and contains diverse persona inferences less likely to be generated by LLMs.
%\antoine{can we say something more positive about KG extraction? Because this just reads as though there was no purpose to this approach vs. the LM-based approach?}
%Within the extrinsic feature, \textit{Routine or Habit} relations are more often presented compared to the relation pairs that connote future and past tenses.
%In detail, by using a highly-handcrafted knowledge graph, the extractive approach results in imbalanced persona knowledge with a high number of \textit{routine or habit} relation labels (i.e., 70\% of total extractive labels) compared to the other main relations.
%But the generative approach, on the other hand, generates a balanced number of main labels.
%This implies that the persona knowledge in natural text learned by InstructGPT-3 could contribute to generating diverse relations.

\paragraph{Persona Interconnectivity}
In addition to containing diverse knowledge from multiple sources, \ourkg{} also contains interesting interconnections among personas, which potentially indicate engaging points of common ground for characters of narratives.
For example, as shown in Figure~\ref{fig:persona_illustrate}, a professional singer’s experience of \textit{studying music at college} is also the routine of a music-major student, which shows a common topic for these two persona to discuss.
Among 40,665 distinctive attributes in \ourkg{}, we find that 9,242 attributes are connected to two or more personas, forming 239,812 bridges, \ie{}, pairs of personas connected via a shared common attribute.\footnote{The number of bridges grows combinatorially with the number of personas sharing an attribute.}
%These bridges demonstrate that \ourkg{} contains rich interconnected world persona knowledge, enabling complex persona reasoning and engaging conver modeling.

\subsection{Attribute Disagreements}
\label{sec:disagreement}
One of our innovations in this work is to introduce InstructGPT-3 as a third annotator to resolve disagreements between human annotators via majority voting. We analyze the disagreements between workers across the annotations as in Table~\ref{tab:final_disagree}, %We track the disagreements between workers when InstructGPT-3 agrees with one of the two workers, or none of InstructGPT-3 and two workers agrees.
and observe that labels from InstructGPT-3 effectively solve many disagreements between human workers.
For the main dimension labeling, $\sim$73\% of the disagreements are solved by adding InstructGPT-3 as a third annotator. However, $\sim$27\% of labels remain \textit{Controversial} when both annotators and GPT3 all disagree in different ways.
%We exclude these persona inferences from the final version of \ourkg{}.
These controversial labels enable further research on the ambiguities in real-world persona types and the potential stereotypes in persona judgments.
%We also find that employing InstructGPT-3 could help to solve the disagreements in labeling \textit{Distinctive} \vs{} \textit{Generic}, although in the LM-based persona collection approach, InstructGPT-3 only generates the distinctive aspects of \textit{head} personas.
%In the interactivity dimension, $\sim$88\% of the disagreements for \textit{Relationship} \vs{} \textit{Self} are left unresolved,\footnote{Most of these disagreements arise from the KG-extracted examples that were re-classified by InstructGPT-3 as corresponding to a \textit{main} dimension: characteristic, routine, goal, experience. As a result, there is no InstructGPT-3 vote for the interactively dimension to resolve the worker disagreement.} but we retain these facts as they may still correspond to good persona inferences along their main dimension.
% However, these unsolved disagreements only occupy $\sim$18\% of the total persona facts.
In the interactivity and distinctiveness dimensions where the labeling schema is binary, disagreements of workers are fully solved by the majority voting with InstructGPT-3, though ambiguous cases may still remain.

\begin{table}[t]
\centering
\resizebox{1.0\columnwidth}{!}{
\smallskip\begin{tabular}
{@{~}lc@{~~~}cc@{~~~}cc@{~~~}c@{~}}
\toprule
\multirow{2}*{\textbf{Dimension}} & \multicolumn{2}{c}{\textbf{GPT3 \& W1/2}} & \multicolumn{2}{c}{\textbf{W1 \& W2}} & \multicolumn{2}{c}{\textbf{All}} \\
\cmidrule(lr){2-3} \cmidrule(lr){4-5} \cmidrule(lr){6-7}
& \textbf{Acc.} & \textbf{F1} & \textbf{Acc.} & \textbf{F1} & \textbf{Acc.} & \textbf{F1} \\
\toprule
Main & 0.854 & 0.851 & 0.872 & 0.810 & 0.857 & 0.845\\
\midrule
Interactivity & 0.907 & 0.844 & 0.924 & 0.837 & 0.913 & 0.842\\
\midrule
Distinctiveness & 0.853 & 0.906 & 0.847 & 0.912 & 0.851 & 0.907\\
\bottomrule
\end{tabular}
}
\caption{Expert evaluation of majority voting quality. \textbf{GPT3 \& W1/2}: InstructGPT-3 agrees with one of the workers and not with the other, \textbf{W1 \& W2}: Two workers agree with each other but not with InstructGPT-3. \textbf{F1} denotes Macro-F1 scores for the main dimension, and F1 scores on the \textit{Relationship} and \textit{Distinctive} classes.}% for the interactivity and distinctiveness dimensions.}
\label{tab:final_accuracy}
\end{table}

% Symbolic analysis - quality (Beatriz):
\paragraph{Expert Study}
%One question that naturally arises when employing InstructGPT-3 as a third annotator in disagreements, and more generally when employing a majority voting approach, is whether this classification decision remains accurate.
However, one question that naturally arises, when employing a majority voting with InstructGPT-3 in the loop, is whether this classification decision remains accurate.
To evaluate this, two experts from our research group manually re-annotate the relations of 825 persona facts in \ourkg{}, and then compare their annotations to the majority voting results to check the voting accuracy.
The 825 persona facts consist of 275 samples from each of the three \ourkg{} subsets where majority voting is employed, that is, when InstructGPT-3 agrees with one of the workers but not the other, and when both workers agree with each other but not with InstructGPT-3. 
% Each subset of the 275 samples further consists of 25 samples from each of the 11 cases where InstructGPT-3 provides different labels in different persona collection approaches, including the five \ourkg{} relation labels in both KG-based and LM-based approaches, and the \textit{No} label in only the KG-based approach.
% To ensure highly precise and reliable expert labels, we also include a third expert to re-check the annotations of the two experts and resolve the few (7\%) disagreement cases between the experts.
%
Experts are required to pass a qualification test by performing 20 test annotations correctly. Furthermore, in the case of disagreements (7\% of cases), a third expert re-checked the annotations of the two experts and resolved the disagreement cases.\footnote{To ensure fairness, the experts do not see the relation labels predicted by crowdworkers and InstructGPT-3.}
%The overall agreement rate of the two experts is 0.93, indicating a very high degree of agreement in our relation re-annotation. For the few disagreement cases between the experts, a third expert was included to decide the final annotation labels.

% \begin{enumerate}
% \item \textbf{Main} -- classifying the entry with the appropriate label, between \textit{Characteristic}, \textit{Routine/Habit}, \textit{Goal/Plan}, and \textit{Experience}
% \item \textbf{Interactivity} -- classifying the entry as either \textit{Him/Herself} or \textit{Relationship}
% \item \textbf{Distinctiveness} -- classifying the entry as either \textit{Generic} or \textit{Distinctive}
%\end{enumerate}
% TODO - do I need to explain each of these categories further (or are they explained in a previous section)?
% silin - no, you don't need, they're explained in the previous section

Table \ref{tab:final_accuracy} presents the accuracy and F1 of the majority voting results, compared to the re-annotations from experts as ground truth labels.
We stratify the results into two cases: the two workers disagree with each other but InstructGPT-3 agrees with one of them, and both workers agree with each other but not with InstructGPT-3.
We observe a high agreement between the experts and the majority vote, with an average accuracy and F1 of 0.874 and 0.865, respectively.
These results validate majority voting with InstructGPT-3 in the loop, showing that InstructGPT-3 serves as a reliable third annotator when disagreements arise.
Moreover, the integration of InstructGPT-3 in the verification loop costs less in terms of time and money compared to adding more human annotators.
%Moreover, the results demonstrate the two instances are directly comparable, establishing InstructGPT-3 as a reliable third annotation worker. This validates our majority voting approach, as it has been shown accurate.
%In addition, this is an extremely efficient approach, as it allows the disambiguation of more than half of all disagreements without introducing any further human labor.
%As such, the introduction of InstructGPT-3 as a third annotator leads to comparable results, while simultaneously having lower temporal and financial associated costs.

However, we note that InstructGPT-3 is not a panacea on its own. While the model effectively resolves worker disagreements, 
%the relation classification of \ourkg{} facts is still challenging when done without human effort. In the main dimension of relation labeling, 
we find that its individual predictions are only correct with $\sim$60\% macro-F1, which is far from the $\sim$85\% macro-F1 with majority voting, indicating that not all \ourkg{} persona relations are known by large-scale language models, and that human crowdsourcing is still necessary to ensure data quality.

\begin{table}[t]
\centering
\resizebox{1.0\columnwidth}{!}{
\smallskip
\begin{tabular}{@{~}l@{~~}c@{~~}c@{~~}c@{~~}c@{~}}
\toprule
              & \textbf{BLEU}  & \textbf{ROUGE-L} & \textbf{METEOR} & \textbf{SkipThoughts} \\
\midrule
GPT-3 (5-shot)     & 71.26          & 72.95            & 50.78           & 68.49     \\
GPT-3.5 (0-shot)   & 57.90          & 63.99            & 47.62           & 61.85     \\
\midrule
\comet-BART        & \textbf{78.04} & \textbf{79.61}   & \textbf{58.88}  & \textbf{75.84}      \\
\bottomrule
\end{tabular}
}
\caption{Automatic evaluation results of \textit{attribute} generation on \ourkg{} test set.}
\label{tab:nlg_results}
\end{table}

\begin{table}[t]
\centering
\resizebox{1.0\columnwidth}{!}{
\smallskip
\begin{tabular}{@{~}l@{~~~}c@{~~~}c@{~~~}c@{~}}
\toprule
            & \textbf{Accept (\%)}  & \textbf{Reject (\%)} & \textbf{No Judgement (\%)} \\
\midrule
GPT-3 (5-shot)   & 96.20          & 3.47            & 0.33  \\
GPT-3.5 (0-shot) & 87.76          & 10.83            & 1.42  \\
\midrule
\comet-BART      & \textbf{97.03} & \textbf{2.94}   & 0.03  \\
\bottomrule
\end{tabular}
}
\caption{Human evaluation results of \textit{attribute} generation on \ourkg{} test set. Crowdworkers judge each fact as \textit{always or likely true} (Accept), \textit{farfetched or invalid} (Reject), or \textit{too unfamiliar to judge} (No Judgment).}
\label{tab:human_results}
\end{table}

\begin{table*}[t]
\centering
\resizebox{1.0\textwidth}{!}{
\smallskip\begin{tabular}{lcccccccc}
\toprule
\multirow{2}*{\textbf{Model}} & \multicolumn{4}{c}{\textbf{Original \personachat{} Profiles}} & \multicolumn{4}{c}{\textbf{Revised \personachat{} Profiles}} \\
\cmidrule(lr){2-5} \cmidrule(lr){6-9}
 & \textbf{PPL} & \textbf{Hits@1 (\%)} & \textbf{F1 (\%)} & \textbf{BLEU (\%)} & \textbf{PPL} & \textbf{Hits@1 (\%)} & \textbf{F1 (\%)} & \textbf{BLEU (\%)} \\
\toprule
\psqbot{}                         &  15.23  &  82.2  & \textbf{19.79} &  0.91  &  18.71  &  68.8  & \textbf{18.92} &  0.71  \\
\psqbot{} + \atomicTT{}          &  15.18  &  81.9  &  18.54  &  0.94  &  18.49  &  72.9  &  17.82  &  0.70  \\
\psqbot{} + \ourkg{}             & \textbf{14.46} & \textbf{83.3} &  19.63 & \textbf{1.02} & \textbf{18.25} & \textbf{75.7} &  18.71  & \textbf{0.75}  \\
\bottomrule
\end{tabular}
}
\caption{Downstream dialogue response generation results on the ConvAI2 \personachat{} dataset. All the results are evaluated on the development set since the test set is not publicly available. We use the trained model provided by \psqbot{} paper to reproduce the baseline results under the same environment as for developing \psqbot{} + \ourkg{}.}
\label{tab:downstream_results_auto}
\end{table*}

\begin{table*}[t]
\centering
\resizebox{1.0\textwidth}{!}{
\smallskip\begin{tabular}{lcccccccc}
\toprule
\multirow{2}*{\textbf{Compared Model}} & \multicolumn{2}{c}{\textbf{Fluency}} & \multicolumn{2}{c}{\textbf{Consistency}} & \multicolumn{2}{c}{\textbf{Engagement}} & \multicolumn{2}{c}{\textbf{Persona Expression}}\\
\cmidrule(lr){2-3} \cmidrule(lr){4-5} \cmidrule(lr){6-7} \cmidrule(lr){8-9}
 & \textbf{win (\%)} & \textbf{lose (\%)} & \textbf{win (\%)} & \textbf{lose (\%)} & \textbf{win (\%)} & \textbf{lose (\%)} & \textbf{win (\%)} & \textbf{lose (\%)} \\
\toprule
\psqbot{}                &      40.0   &    5.5     &     54.0      &     22.5      &      48.5     &      28.5     &      57.0     &     25.5    \\
\psqbot{} + \atomicTT{} &      17.5   &    4.5     &     37.5      &     24.5      &      46.5     &      22.0     &      57.5     &     20.0    \\
Human                    &      5.0    &    6.0     &     20.0      &     43.5      &      25.0     &      40.0     &      21.5     &     35.0    \\
\bottomrule
\end{tabular}
}
\caption{Pairwise comparisons of dialogue response generation between \psqbot{} + \ourkg{} versus other baseline models. \textbf{Human} denotes the comparison with gold responses. Ties are not shown.}
\label{tab:downstream_results_human}
\end{table*}

\section{Generalizing Persona Knowledge} 
% or Neural KG Comparison
Following the neural KG analysis method proposed by \citealp{hwang2021comet}, we assess whether \ourkg{} could be used to train inference generators that hypothesize persona knowledge.
We train a BART-based \citep{lewis2020bart} \comet{} \citep{bosselut2019comet} knowledge generator (\comet-BART) based on a held-out training set ($\sim$65K facts) of \ourkg{}, where the model learns to generate the \textit{tail} attribute of a fact given its \textit{head} persona and relation.
We evaluate \comet-BART on a test set from \ourkg{} containing 3030 facts with unique \textit{head}-relation combinations. As baselines, we compare to a few-shot GPT-3 \citep{brown2020language} that uses 5 randomly sampled training facts (with same relation as the testing fact) to prompt the \textit{tail} knowledge generation and a zero-shot GPT-3.5 (text-davinci-003) baseline model. These baselines compare \ourkg{} training to larger LMs that use both in context-learning and instruction tuning. %without in-context \ourkg{} examples.
We conduct both automatic and human evaluations on the knowledge generators, with results shown in Tables~\ref{tab:nlg_results} and \ref{tab:human_results}.\footnote{We include more implementation details of our neural KG analysis in Appendix~\ref{sec:neural_analysis}.}
 
Compared to few-shot GPT-3, \comet-BART trained on \ourkg{} achieves overall better automatic evaluation results on various NLG metrics, despite being a much smaller model.\footnote{GPT-3 and \comet-BART have 175B and 440M parameters, respectively.}
In the human evaluation, we find that facts generated by \comet-BART receive a high acceptance rate by crowdworkers for plausibility, slightly beating few-shot GPT-3. We also find that zero-shot GPT-3.5 model, although more advanced than the GPT-3 baseline model, scores, on average, $\sim$15.3\% and $\sim$9.3\% lower than \comet-BART in terms of automatic metrics and human acceptance, respectively. 
% This shows that large-scale LMs need curated few-shot \ourkg{} examples to generate decent persona knowledge.
All above results indicate that \ourkg{} can serve as a reliable persona knowledge base, which enables light-weight LMs to learn knowledge generation capabilities comparable to large-scale LMs.

\section{Enhancing Dialogue Systems}
\label{dowstream_dialogue}
As our knowledge graph \ourkg{} covers rich world persona knowledge, we validate whether acccess to this knowledge enables better persona modeling in downstream narrative systems.
Using \ourkg{}, we augment a persona-grounded dialogue model \psqbot{} \cite{liu2020you} developed on the ConvAI2 \citep{dinan2020second} \personachat{} \citep{zhang2018personalizing} dataset.
We link facts from \ourkg{} to \personachat{} dialogues, thereby extending \psqbot{}'s persona perception and augmenting its dialogue response generation.\footnote{Downstream application details are in Appendix~\ref{sec:appendix_downstream}.}

We evaluate our models based on both original and revised interlocutor profiles provided in the ConvAI2 \personachat{} dataset, and measure the perplexity (\textbf{PPL}), word-level \textbf{F1}, and cumulative 4-gram \textbf{BLEU} \citep{papineni2002bleu} of the generated responses compared to the references.
We also follow ConvAI2 to measure \textbf{Hits@1}, \ie{}, the probability that real response is ranked the highest by the model among 20 candidates.

\paragraph{Persona Knowledge Linking}
We link \ourkg{} knowledge to interlocutors based on both their \personachat{} profiles and their utterances in the dialogue.
For each interlocutor, we extract all statements in their profile, as well as first-person sentences in their utterances.
Then, we follow a commonsense fact linking benchmark, \comfact{} \citep{gao2022comfact}, to link relevant facts from \ourkg{} to each extracted statement or sentence.
We remove linked facts that are labeled as \textit{Generic} in the distinctiveness dimension, \ie{}, have little effect on distinguishing this persona from others.

For each interlocutor, we randomly sample 5 \ourkg{} facts that are linked to their \personachat{} profile,\footnote{Due to the model capacity limitation of the baseline \psqbot{}, we only sample a subset of linked \ourkg{} facts as the extended persona knowledge for each interlocutor.} and convert them into natural language statements to form their extended persona knowledge.\footnote{Fact preprocessing details are in Appendix~\ref{sec:neural_analysis} and \ref{sec:appendix_downstream}.}
Our augmented model is denoted as \psqbot{} + \ourkg{}.
To compare \ourkg{}'s persona-centric knowledge augmentations with general commonsense augmentations, we also evaluate another baseline model \psqbot{} + \atomicTT{}, where we follow \citealp{majumder2020like} to extend interlocutor personas with 5 randomly sampled commonsense inferences from the \comet{}-\atomicTT{} model \citep{hwang2021comet}.

\paragraph{Results}
In Table~\ref{tab:downstream_results_auto}, we show that \psqbot{} + \ourkg{} significantly outperforms \psqbot{} on PPL and Hits@1,\footnote{significant at $p$<$0.02$ and $p$<$0.01$, respectively, in paired sample t-test} 
% p-value 0.011 and 0.008 in paired sample t-test), 
and has comparable F1 and BLEU scores.
Compared to \psqbot + \atomicTT, \psqbot{} + \ourkg{} also demonstrates a clear improvement across all metrics, indicating the importance of augmenting narrative systems with persona-grounded commonsense knowledge.
% These results hint at \ourkg{}'s great potential to benefit narrative systems by enabling access to better world persona knowledge, which is irreplaceable by general commonsense inferences.

\paragraph{Human Evaluation}
Automatic metrics are not fully reliable for evaluating dialogue systems \citep{liu2016not,novikova2017we}, so we also conduct human evaluations on the dialogue responses.
We make pairwise comparisons between \psqbot{} + \ourkg{} and other baseline models, based on their generated responses to 200 randomly sampled dialogue histories (100 each with original and revised \personachat{} profiles). Two expert annotators from our research group manually compare four aspects of the response generation quality: \textbf{fluency}, whether the response is fluent and understandable, \textbf{consistency}, where the response is consistent with the dialogue history, \textbf{engagement}, whether the response is engaging and interesting, and \textbf{persona expression}, whether the response demonstrates persona information related to the interlocutor's profile.
To ensure the fairness and reliability of our human evaluation, similar to Sec.~\ref{sec:disagreement}, we require each expert to pass a qualification test on 10 pairwise comparisons, and also include a third qualified expert to re-check the evaluation results. We note that both expert annotators do not see the source model from which each response is generated.
%\footnote{\textcolor{red}{We showcase our interface for the expert qualification test in Appendix~\ref{sec:appendix_downstream}.}}

The human evaluation results in Table~\ref{tab:downstream_results_human} show that \psqbot{} + \ourkg{} generates more consistent and engaging dialogues compared to other neural baselines, demonstrating that persona commonsense knowledge is a key contributor to the conversation consistency and engagement.
However, \psqbot{} + \ourkg{} still has room for improvement compared to human performance.
%This reveals the valuable research direction of developing better world persona modeling in narratives.

\begin{table}[t]
\centering
\resizebox{1.0\columnwidth}{!}{
\smallskip\begin{tabular}{@{~}c@{~~}ccccc@{~}}
\toprule
\multirow{2}*{\textbf{\#CA}} & \multirow{2}*{\textbf{\#DR}} & \multicolumn{2}{c}{\textbf{Consistency}} & \multicolumn{2}{c}{\textbf{Engagement}} \\
\cmidrule(lr){3-4} \cmidrule(lr){5-6}
& & \textbf{win (\%)} & \textbf{lose (\%)} & \textbf{win (\%)} & \textbf{lose (\%)} \\
\toprule
 0     &  59  &   42.4   &    23.7   &   44.1    &   28.8   \\
 1     &  45  &   57.8   &    24.4   &   44.4    &   24.4   \\
 $>$ 1 &  96  &   59.3   &    20.8   &   53.1    &   30.2   \\
\bottomrule
\end{tabular}
}
\caption{Pairwise comparisons of dialogue response generation between \psqbot{} + \ourkg{} versus \psqbot{}, stratified by the number of shared \ourkg{} attributes between interlocutors. ``\#CA'' denotes the number of common attributes shared by the two interlocutors’ linked \ourkg{} knowledge. ``\#DR'' denotes the number of dialogue responses evaluated in each stratified experiment. Ties are not shown.}
\label{tab:downstream_results_human_fine}
\end{table}

% \textcolor{red}{To investigate how \ourkg{}'s interconnected knowledge improves the consistency and engagement of dialogues, we stratify the pairwise comparison between \psqbot{} + \ourkg{} versus \psqbot{} according to the overlap of two speakers' linked \ourkg{} knowledge.
% As shown in Table~\ref{tab:downstream_results_human_fine}, we stratify the comparison on dialogues where the two speakers' linked \ourkg{} personas have 0, 1 or more than 1 attributes in common.
% We find that the winning rates of P2Bot w/ PeaCoK on dialogue consistency and engagement increase as the overlap of the two speakers' linked \ourkg{} personas becomes larger.
% This demonstrates that more connections between interlocutors leads to more consistent and engaging conversations, which highlights the importance of learning interconnected world persona knowledge in narratives.}

Perhaps most interestingly, though, we find that \ourkg{}'s impact on the consistency and engagement of dialogues is most pronounced when there are interconnections between the personas of the interlocutors. We stratify the pairwise comparison between \psqbot{} + \ourkg{} versus \psqbot{} from Table~\ref{tab:downstream_results_human} based on the overlap of the two interlocutors' linked \ourkg{} knowledge.
In Table~\ref{tab:downstream_results_human_fine}, we show the results of this stratification across the cases where the interlocutors have 0, 1 or more than 1 shared attributes.
Specifically, we find that the winning rates of \psqbot{} w/ \ourkg{} on dialogue consistency and engagement increase as the overlap of the two speakers' linked \ourkg{} personas becomes larger, demonstrating that more connections between interlocutors leads to more consistent and engaging conversations, and highlighting the importance of learning interconnected world persona knowledge in narratives.

\section{Conclusion}
In this work, we propose a persona commonsense knowledge graph, \ourkg{}, to complement the real-world picture of personas that ground consistent and engaging narratives.
\ourkg{} consists of $\sim$100K persona commonsense inferences, distilled from existing KGs and pretrained LMs, across five dimensions of persona knowledge identified in prior literature on human interactive behaviours. 
Our analysis and experiments demonstrate that \ourkg{} contains high-quality inferences whose connectivity provides many instances of common ground between personas, improving the consistency and engagement of downstream narrative systems.

\section*{Limitations}
We acknowledge a few limitations in this work. First, \ourkg{} cannot be comprehensive. Persona knowledge is very broad and our resource cannot cover all dimensions of personas, nor all attributes of these dimensions. We select five dimensions of personas that we found salient from background literature in human interaction, and we distill attributes for these dimensions from \atomicTT{}, \comet{} and InstructGPT-3. These resources, while rich in knowledge, only represent a subset of possible background resources for the construction of \ourkg (among other KGs and pretrained language models). Furthermore, the primary language of these three resources is English, making \ourkg{} a solely English resource.
%Our knowledge base also cannot have an exhaustive coverage of the target (persona commonsense) knowledge, due to its limitless amount and boundless range in reality.
% However, we use diverse knowledge collection methods to gather the most common aspects of our target knowledge in a rich amount.
Finally, in downstream narrative experiments, the usage of our augmented persona knowledge is constrained by the capacity of baseline model, which leaves for future work the exploration of downstream persona knowledge augmentation on a larger scale.

\section*{Ethics Statement}
Our work is approved by our institution's human research ethics committee to conduct human-centric or ethics-related experiments, \eg{}, crowdsourcing and human evaluations. Topic-wise, our research develops a knowledge graph of commonsense knowledge about personas to augment understanding of characters and their interactions in diverse narratives. Given that some of the attributes are extracted from previous KGs or generated by LMs, we cannot guarantee our knowledge graph does not contain attribute alignments with negative connotations that could provide undesired information to a downstream system. However, we took the following steps to mitigate this effect. First, the set of personas we include in \ourkg{} was manually filtered to not include stereotypical and harmful roles, thereby limiting the negative associations of the personas themselves. Second, we explicitly prompted the LM to generate optimistic attributes about personas, which has been shown in prior work to reduce the toxicity of outputs \citep{Schick2021debiasing}. Finally, each attribute in \ourkg{} is explicitly validated by two human workers for toxicity, providing a final opportunity for workers to flag problematic content. However, we acknowledge that none of these safeguards are perfect, as language models may still produce toxic outputs and annotators may have differing opinions on what constitutes toxic content \citep{sap2022annotatorsWithAttitudes}.  %, which we did notice had a helpful effect in reducing negative outputs. 
% Scientific work published at ACL 2023 must comply with the ACL Ethics Policy.\footnote{\url{https://www.aclweb.org/portal/content/acl-code-ethics}} We encourage all authors to include an explicit ethics statement on the broader impact of the work, or other ethical considerations after the conclusion but before the references. The ethics statement will not count toward the page limit (8 pages for long, 4 pages for short papers).

% \section*{Acknowledgements}
% Thanks...

\section*{Acknowledgements}
We thank Gail Weiss, Syrielle Montariol, Graciana Aad and Mete Ismayil for reading and providing comments on drafts of this paper. We also gratefully acknowledge the support of Innosuisse under PFFS-21-29, the EPFL Science Seed Fund, the EPFL Center for Imaging, Sony Group Corporation, and the Allen Institute for AI. 

% Entries for the entire Anthology, followed by custom entries
\bibliography{main}
\bibliographystyle{acl_natbib}

\appendix

\section{\ourkg{} Construction Details}
\label{sec:appendix_construction}

\paragraph{Head Persona Selection}
Table~\ref{tab:prompt_filtering} shows our designed prompt for InstructGPT-3 \textit{head} persona filtering described in Sec.~\ref{sec:head_selection}.
We preprocess our extracted human and event-based entities to make them fit into the prompt.
Specifically, we fill each human entity into the template ``I am a(n) \_\_\_.'' to convert it into a natural language sentence.
We also replace the general token ``PersonX'' in each even-based entity with the pronoun ``I'', and lemmatize the third person singular in its verbs.
To build the integral statement (final \textit{head} persona in \ourkg{}) that combines a human entity with each of its derived event-based entity, we instead replace the even-based entity's ``PersonX'' token with ``who'', and then append it to the converted sentence of its human entity.
Note that for each human entity itself or event-based entity that contains a human entity (\ie{}, the first type of derived event-based entities), we directly include its converted sentence alone as one of the \textit{head} persona statements in \ourkg{}.

\paragraph{KG-Based Tail Attribute Collection}
We use \atomicTT{} as the background resource for KG-based \textit{tail} attribute collection described in Sec.~\ref{sec:tail_selection}.
This advanced KG contains 1.33M general social commonsense inferences based on a rich variety of entities, including 0.21M inferences about physical objects, 0.20M inferences centered on daily events, and other 0.92M inferences based on social interactions.
Table~\ref{tab:relation_atomic} lists the 10 \atomicTT{} relations that we consider as potentially related to persona knowledge, which we use to query \textit{tail} attributes from \atomicTT{} KG and \comet{}, based on each original entity collected in the \textit{head} persona selection (Sec.~\ref{sec:head_selection}).

\begin{table}[t]
\centering
\resizebox{1.0\columnwidth}{!}{
\smallskip\begin{tabular}{@{~}ll@{~}}
\toprule
\multicolumn{2}{l}{Does the phrase distinctively entail the role of the person in the script?} \\
\midrule
Script: I am an actor.           & Script: I am a secretary.         \\
Phrase: I am a movie star.       & Phrase: I write official documents. \\
Answer: Yes                      & Answer: Yes                         \\
 & \\
Script: I am an actor.           & Script: I am a secretary.         \\
Phrase: I sing a song.           & Phrase: I have a job interview coming up. \\
Answer: No                       & Answer: No                          \\
 & \\
Script: I am an accountant.      & Script: I am a conductor.          \\
Phrase: I have a CPA license.    & Phrase: I unite performers in an orchestra.\\
Answer: Yes                      & Answer: Yes                         \\
 & \\
Script: I am an accountant.      & Script: I am a conductor.          \\
Phrase: I work as a cashier.     & Phrase: I want to play an instrument. \\
Answer: No                       & Answer: No                          \\
 & \\
Script: I am a student.          & Script: I am a curator.           \\
Phrase: I finish my degree.      & Phrase: I manage the exhibition. \\
Answer: Yes                      & Answer: Yes                         \\
 & \\
Script: I am a student.          & Script: I am a curator.           \\
Phrase: I make a pot of coffee. & Phrase: I work with animals.    \\
Answer: No                       & Answer: No                          \\
 & \\
Script: I am a runner.           & Script: I am a thrifty person.     \\
Phrase: I run a marathon.        & Phrase: I want to save money.     \\
Answer: Yes                      & Answer: Yes                         \\
 & \\
Script: I am a runner.           & Script: I am a thrifty person.     \\
Phrase: I run across the street. & Phrase: I love shopping.        \\
Answer: No                       & Answer: No                          \\
\bottomrule
\end{tabular}
}
\caption{Instruction and in-context examples used for InstructGPT-3 \textit{head} persona filtering.}
\label{tab:prompt_filtering}
\end{table}

\begin{table}[t]
\centering
\resizebox{1.0\columnwidth}{!}{
\smallskip\begin{tabular}{ll}
\toprule
\textbf{Relation}  & \textbf{Relation Description} \\
\toprule
HasProperty & the person is characterized by being/having \\
CapableOf   & the person is capable of \\
Desires     & the person desires \\
xNeed       & but before, the person needs \\
xAttr       & the person is seen as \\
xEffect     & as a result, the person will \\
xReact      & as a result, the person feels \\
xWant       & as a result, the person wants \\
xIntent     & because the person wants \\  
\bottomrule
\end{tabular}
}
\caption{Commonsense relations in \atomicTT{} which are potentially related to personas.}
\label{tab:relation_atomic}
\end{table}

\begin{table*}[t]
\centering
\resizebox{1.0\textwidth}{!}{
\smallskip\begin{tabular}{lll}
\toprule
\textbf{Characteristic} & \textbf{Routine or Habit} & \textbf{Goal or Plan} \\
\midrule
Guess a character trait of the person in  the clue, & Guess what the person in the clue regularly or consistently & Guess what the person in the clue will do or achieve in \\
which can distinguish this  person from others.     & does, which can distinguish this person from others.        & the future, which can distinguish this person from others.\\
\midrule
\multicolumn{3}{c}{\textbf{Simple Head Personas}} \\
\midrule
Clue: I become an accountant. & Clue: I become an accountant.                  & Clue: I become an accountant. \\
Characteristic: good at math    & Routine or Habit: analyze financial information & Goal or Plan: to have my own audit firm \\
& & \\
Clue: I want to be an actor.           & Clue: I want to be an actor.         & Clue: I want to be an actor. \\
Characteristic: interested in performing & Routine or Habit: take acting classes & Goal or Plan: to get auditions \\
& & \\
Clue: I am an alert person.        & Clue: I am an alert person.          & Clue: I am an alert person. \\
Characteristic: sensitive to danger & Routine or Habit: do reconnaissance & Goal or Plan: to keep his children safe \\
& & \\
Clue: I work as a lion tamer. & Clue: I work as a lion tamer. & Clue: I work as a lion tamer. \\
Characteristic: animal lover    & Routine or Habit: train lions  & Goal or Plan: to put on a lion show \\
& & \\
Clue: I am a successful store owner.     & Clue: I am a successful store owner. & Clue: I am a successful store owner. \\
Characteristic: excellent business acumen & Routine or Habit: manage inventory   & Goal or Plan: to open another store location \\
\midrule
\multicolumn{3}{c}{\textbf{Complex Head Personas}} \\
\midrule
Clue: I am an accountant who have a CPA license. & Clue: I am an accountant who have a CPA license. & Clue: I am an accountant who have a CPA license.\\
Characteristic: good at interpreting financial records  & Routine or Habit: prepare financial reports & Goal or Plan: to increase company profits \\
& & \\
Clue: I am an actor who is a movie star.  & Clue: I am an actor who is a movie star.   & Clue: I am an actor who is a movie star. \\
Characteristic: devoted in acting career  & Routine or Habit: participate in film shoots & Goal or Plan: to win a Grammy award \\
& & \\
Clue: I am a successful store owner who have many customers. & Clue: I am a successful store owner who have many customers. & Clue: I am a successful store owner who have many customers.\\
Characteristic: have a customer-centric way of thinking  & Routine or Habit: control the purchase of goods & Goal or Plan: to reach new target customers \\
& & \\
Clue: I am an alert person who is observant. & Clue: I am an alert person who is observant.  & Clue: I am an alert person who is observant. \\
Characteristic: sensitive to hidden danger    & Routine or Habit: pay attention to surroundings & Goal or Plan: to uncover potential hazards \\
& & \\
Clue: I am a lion tamer who love animals. & Clue: I am a lion tamer who love animals.  & Clue: I am a lion tamer who love animals. \\
Characteristic: calm with facing lions    & Routine or Habit: take good care of lions & Goal or Plan: to put on a lion shows \\
\bottomrule
\end{tabular}
}
\caption{Instructions and in-context examples used for InstructGPT-3 \textit{tail} attribute generation with respect to the \textit{Characteristic}, \textit{Routine or Habit} and \textit{Goal or Plan} relations.}
\label{tab:prompt_generation_1}
\end{table*}

\begin{table*}[t]
\centering
\resizebox{0.8\textwidth}{!}{
\smallskip\begin{tabular}{ll}
\toprule
\textbf{Experience} & \textbf{Relationship} \\
\toprule
Guess what the person in the clue did in the past, & Guess a relationship that the person in the clue has with other people \\
which can distinguish this person from others.     & or social groups, which can distinguish this person from others. \\
\midrule
\multicolumn{2}{c}{\textbf{Simple Head Personas}} \\
\midrule
Clue: I become an accountant.      & Clue: I become an accountant.  \\
Experience: got a degree in finance  & Relationship: work with clients \\
& \\
Clue: I want to be an actor.    & Clue: I want to be an actor.             \\
Experience: auditioned for a play & Relationship: sign up with a film company \\
& \\
Clue: I am an alert person.             & Clue: I am an alert person.          \\
Experience: discovered a security breach & Relationship: keep his friends safe \\
& \\
Clue: I work as a lion tamer.            & Clue: I work as a lion tamer. \\
Experience: qualified as an animal trainer & Relationship: supervised by the zoo director \\
& \\
Clue: I am a successful store owner.              & Clue: I am a successful store owner. \\
Experience: studied business management in college & Relationship: attract customers with promotions \\
\midrule
\multicolumn{2}{c}{\textbf{Complex Head Personas}} \\
\midrule
Clue: I am an accountant who have a CPA license. & Clue: I am an accountant who have a CPA license. \\
Experience: passed the accounting qualification exam  & Relationship: provide financial information to business owners \\
& \\
Clue: I am an actor who is a movie star.  & Clue: I am an actor who is a movie star. \\
Experience: acted in many good movies  & Relationship: have a stand-in actress \\
& \\
Clue: I am a successful store owner who have many customers. & Clue: I am a successful store owner who have many customers. \\
Experience: received a business license  & Relationship: attract customers with promotions  \\
& \\
Clue: I am an alert person who is observant. & Clue: I am an alert person who is observant.  \\
Experience: discovered a security breach      & Relationship: warned people around about a danger \\
& \\
Clue: I am a lion tamer who love animals. & Clue: I am a lion tamer who love animals.  \\
Experience: qualified as an animal trainer    & Relationship: entertain zoo visitors \\
\bottomrule
\end{tabular}
}
\caption{Instructions and in-context examples used for InstructGPT-3 \textit{tail} attribute generation with respect to the \textit{Experience} and \textit{Relationship} relations.}
\label{tab:prompt_generation_2}
\end{table*}

\paragraph{LM-Based Tail Persona Collection}
Tables \ref{tab:prompt_generation_1} and \ref{tab:prompt_generation_2} show the prompts provided to InstructGPT-3 \textit{tail} to generate attributes for each persona (Sec.~\ref{sec:tail_selection}), based on each converted persona statement derived from the head persona selection (Sec.~\ref{sec:head_selection}).
We use 2 different sets of in-context examples to prompt the InstructGPT-3 generation.
Specifically, examples under the \textbf{Simple Head Personas} block are used for \textit{head} statements converted from human entities or event-based entities that directly contain human entities (the first type of derived event-based entities).
While examples under the \textbf{Complex Head Personas} block are used for event-based entities that do not contain human entities (the second and third types of derived event-based entities), where the event-based entity is combined with its source human entity to form a integral statement.

\begin{figure*}[t]
\centering
\includegraphics[width=1.0\textwidth]{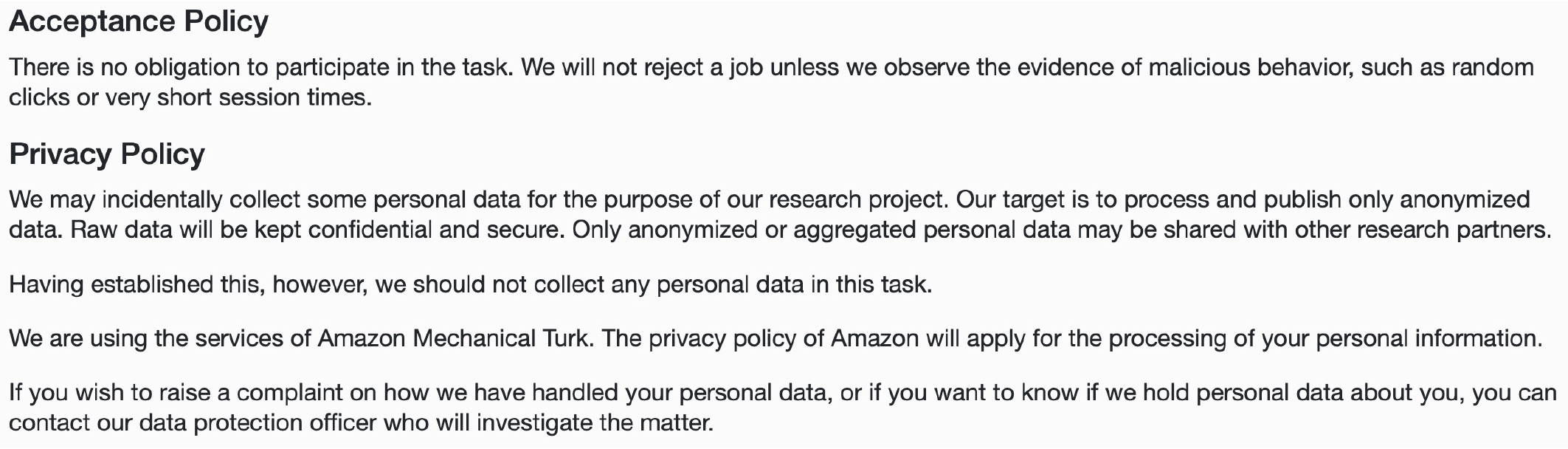}
\caption{Screenshot of our acceptance and privacy policy for crowdsourcing.}
\label{policy_crowd}
\end{figure*}

\begin{figure*}[t]
\centering
\includegraphics[width=1.0\textwidth]{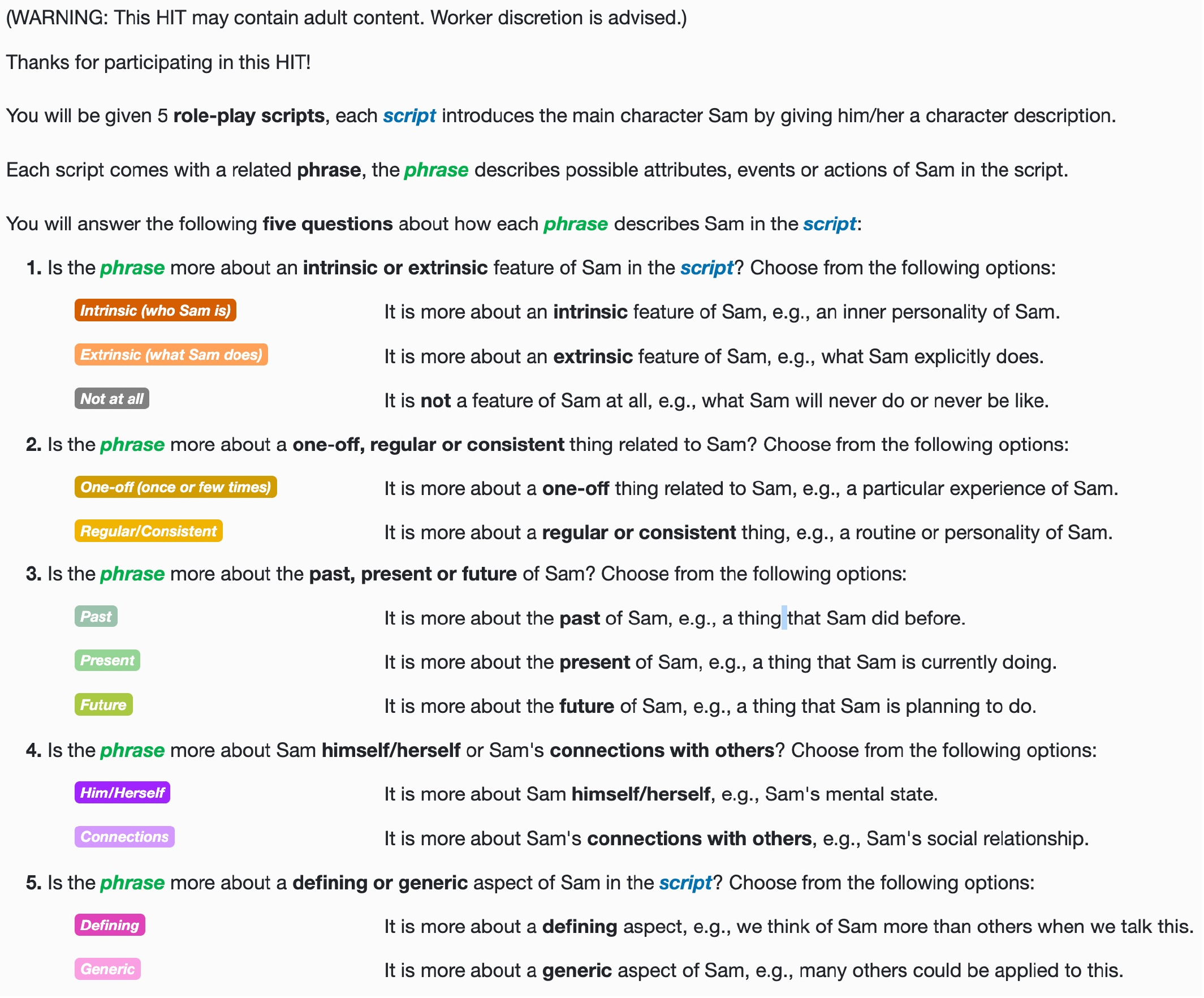}
\caption{Screenshot of our relation classification instruction for crowdsourcing.}
\label{relation_crowd}
\end{figure*}

\paragraph{Crowdsourcing Relation Classification}
We conduct a worker qualification for our persona relation classification described in Sec.~\ref{sec:relation}.
To select native English speakers, we focus on the group of workers whose locations are in the USA.
We test workers with 10 \textit{head} personas, each with 2 \textit{tail} personas (\ie{}, totally 20 \textit{head}-\textit{tail} persona pairs), and select workers who can reasonably annotate 18 or more (\ie{}, $\geq$90\%) relations between the given \textit{head} and \textit{tail} personas.
Finally, 72 out of 207 workers are selected as qualified.
We pay each worker \$0.30 for doing every 5 annotations.
The average hourly wage for each worker is about \$18.00, which is in the acceptable range of hourly wage suggested by Amazon Mechanical Turk.
Figure~\ref{policy_crowd} and \ref{relation_crowd} shows the screenshots of our acceptance policy, privacy policy, and task instruction used for crowdsourcing.

\paragraph{Majority Voting}
Table~\ref{tab:prompt_classification_main}, \ref{tab:prompt_classification_interact} and \ref{tab:prompt_classification_distinct} show the prompts provided to InstructGPT-3 to label relations as the majority vote among worker disagreements (Sec.~\ref{sec:relation}).
Similar to the InstructGPT-3 \textit{tail} attribute generation (Sec.~\ref{sec:tail_selection}), we use 2 different sets of in-context examples to handle the complexity differences in the \textit{head} persona statements. The verbalizers that we use for each labeling class are \textit{characteristic}, \textit{routine}, \textit{plan}, \textit{experience} \& \textit{no} in the main dimension; \textit{relationship} \& \textit{self} in the interactivity dimension; and \textit{distinctive} \& \textit{generic} in the distinctiveness dimension.

% \paragraph{Claim of Usage}
% Our proposed knowledge graph is intended to be used for only research purposes, any usage of our knowledge graph that is outside of research contexts is not allowed.

\begin{table*}[t]
\centering
\resizebox{0.8\textwidth}{!}{
\smallskip\begin{tabular}{ll}
\toprule
\multicolumn{2}{l}{Judge whether the phrase describes a characteristic, a routine, a plan, or an experience of the person in the script.} \\
\midrule
\multicolumn{2}{c}{\textbf{Simple Head Personas}} \\
\midrule
Script: I want to be an actor. & Script: I become a lonely person. \\
Phrase: good at performing       & Phrase: introverted                 \\
Answer: characteristic           & Answer: characteristic              \\
 & \\
Script: I want to be an actor. & Script: I become a lonely person. \\
Phrase: take acting classes     & Phrase: spend time alone           \\
Answer: routine                  & Answer: routine                     \\
 & \\
Script: I want to be an actor. & Script: I become a lonely person. \\
Phrase: get an audition         & Phrase: find a partner             \\
Answer: plan                     & Answer: plan                        \\
 & \\
Script: I want to be an actor. & Script: I become a lonely person. \\
Phrase: enjoy a good play       & Phrase: divorce from wife          \\
Answer: experience               & Answer: experience                  \\
 & \\
Script: I want to be an actor. & Script: I become a lonely person. \\
Phrase: play in a band          & Phrase: jittery                 \\
Answer: no                       & Answer: no                          \\
\midrule
\multicolumn{2}{c}{\textbf{Complex Head Personas}} \\
\midrule
Script: I am an actor who is a movie star. & Script: I am a lonely person who need someone to talk to.\\
Phrase: good at performing                  & Phrase: depressed                 \\
Answer: characteristic                      & Answer: characteristic             \\
 & \\
Script: I am an actor who is a movie star. & Script: I am a lonely person who need someone to talk to.\\
Phrase: attend movie auditions              & Phrase: stay home alone             \\
Answer: routine                             & Answer: routine                     \\
 & \\
Script: I am an actor who is a movie star. & Script: I am a lonely person who need someone to talk to.\\
Phrase: win a Grammy award                  & Phrase: find a friend to speak to   \\
Answer: plan                                & Answer: plan                        \\
 & \\
Script: I am an actor who is a movie star. & Script: I am a lonely person who need someone to talk to.\\
Phrase: have worked in good movies          & Phrase: divorce from wife           \\
Answer: experience                          & Answer: experience                  \\
 & \\
Script: I am an actor who is a movie star. & Script: I am a lonely person who need someone to talk to.\\
Phrase: play in a band                     & Phrase: jittery                      \\
Answer: no                                  & Answer: no                           \\
\bottomrule
\end{tabular}
}
\caption{Instruction and in-context examples used for InstructGPT-3 relation classification in the main dimension.}
\label{tab:prompt_classification_main}
\end{table*}

\begin{table*}[t]
\centering
\resizebox{0.8\textwidth}{!}{
\smallskip\begin{tabular}{ll}
\toprule
\multicolumn{2}{l}{Judge whether the phrase describes a relationship of the person in the script, or just the person himself.} \\
\midrule
\multicolumn{2}{c}{\textbf{Simple Head Personas}} \\
\midrule
Script: I want to be an actor. & Script: I become a lonely person. \\
Phrase: join an acting club    & Phrase: have few friends             \\
Answer: relationship           & Answer: relationship              \\
 & \\
Script: I want to be an actor. & Script: I become a lonely person. \\
Phrase: enjoy a good play     & Phrase: spend time alone           \\
Answer: self                  & Answer: self                     \\
 & \\
Script: I want to be an actor.      & Script: I become a lonely person. \\
Phrase: learn from famous actors     & Phrase: divorce from wife          \\
Answer: relationship                  & Answer: relationship               \\
 & \\
Script: I want to be an actor. & Script: I become a lonely person. \\
Phrase: good at performing       & Phrase: introverted          \\
Answer: self                      & Answer: self                     \\
\midrule
\multicolumn{2}{c}{\textbf{Complex Head Personas}} \\
\midrule
Script: I am an actor who is a movie star. & Script: I am a lonely person who need someone to talk to.\\
Phrase: gain a lot of fans                  & Phrase: have few friends                 \\
Answer: relationship                        & Answer: relationship               \\
 & \\
Script: I am an actor who is a movie star. & Script: I am a lonely person who need someone to talk to.\\
Phrase: good at performing                  & Phrase: stay home alone             \\
Answer: self                                & Answer: self                     \\
 & \\
Script: I am an actor who is a movie star. & Script: I am a lonely person who need someone to talk to.\\
Phrase: sign with a film company              & Phrase: divorce from wife   \\
Answer: relationship                        & Answer: relationship               \\
 & \\
Script: I am an actor who is a movie star. & Script: I am a lonely person who need someone to talk to.\\
Phrase: win a Grammy award                   & Phrase: depressed           \\
Answer: self                                & Answer: self                     \\
\bottomrule
\end{tabular}
}
\caption{Instruction and in-context examples used for InstructGPT-3 relation classification in the interactivity dimension.}
\label{tab:prompt_classification_interact}
\end{table*}

\begin{table*}[t]
\centering
\resizebox{0.8\textwidth}{!}{
\smallskip\begin{tabular}{ll}
\toprule
\multicolumn{2}{l}{Judge whether the phrase describes a distinctive trait of the person in the script, or just a generic aspect of a person.} \\
\midrule
\multicolumn{2}{c}{\textbf{Simple Head Personas}} \\
\midrule
Script: I want to be an actor. & Script: I become a lonely person. \\
Phrase: take acting classes    & Phrase: spend time alone             \\
Answer: distinctive             & Answer: distinctive              \\
 & \\
Script: I want to be an actor. & Script: I become a lonely person. \\
Phrase: make money              & Phrase: go out to a mall           \\
Answer: generic                  & Answer: generic                     \\
 & \\
Script: I want to be an actor.      & Script: I become a lonely person. \\
Phrase: join an acting club          & Phrase: introverted          \\
Answer: distinctive             & Answer: distinctive              \\
 & \\
Script: I want to be an actor. & Script: I become a lonely person. \\
Phrase: hardworking              & Phrase: ask for help          \\
Answer: generic                  & Answer: generic                     \\
\midrule
\multicolumn{2}{c}{\textbf{Complex Head Personas}} \\
\midrule
Script: I am an actor who is a movie star. & Script: I am a lonely person who need someone to talk to.\\
Phrase: gain a lot of fans                  & Phrase: depressed                 \\
Answer: distinctive                         & Answer: distinctive              \\
 & \\
Script: I am an actor who is a movie star. & Script: I am a lonely person who need someone to talk to.\\
Phrase: hardworking                         & Phrase: go out to a mall             \\
Answer: generic                              & Answer: generic                     \\
 & \\
Script: I am an actor who is a movie star. & Script: I am a lonely person who need someone to talk to.\\
Phrase: good at performing                  & Phrase: have few friends   \\
Answer: distinctive                         & Answer: distinctive              \\
 & \\
Script: I am an actor who is a movie star. & Script: I am a lonely person who need someone to talk to.\\
Phrase: make money                          & Phrase: ask for help           \\
Answer: generic                  & Answer: generic                     \\
\bottomrule
\end{tabular}
}
\caption{Instruction and in-context examples used for InstructGPT-3 relation classification in the distinctiveness dimension.}
\label{tab:prompt_classification_distinct}
\end{table*}

\section{\ourkg{} Analysis Details}
\label{sec:appendix_examples}
Table~\ref{tab:fine_grained_stats} shows the fine-grained statistics of persona relations included in \ourkg{}.
Each \ourkg{} fact's relation consists of three dimensions of labels as shown in Figure~\ref{tab:feature_relation}.
The combinations of \textit{Routine or Habit}, \textit{Self} and \textit{Distinctive} labels is the most frequent relation in \ourkg{}, which implies that individual daily activities might be the most common topic involved in human interactions.
Table~\ref{tab:example_persona_full} shows several examples of persona facts in \ourkg{}, which showcases our knowledge graph's rich commonsense inferences on persona-grounded knowledge.
%Nevertheless, some facts in \ourkg{} maintain ambiguous relation labels, \eg{}, \textit{get tips} can be seen as either an individual routine of a \textit{waiter} or his interaction with customers, showing that it is challenging to construct deterministic world persona knowledge.

\begin{table*}[t]
\centering
\resizebox{0.8\textwidth}{!}{
\smallskip\begin{tabular}
{lccccccc}
\toprule
\multirow{2}*{\textbf{Main Label}} & \multicolumn{3}{c}{\textbf{Distinctive}} &\multicolumn{3}{c}{\textbf{Generic}} & \multirow{2}{*}{\textbf{Total}} \\
\cmidrule(lr){2-4}  \cmidrule(lr){5-7}
& \textbf{Relationship} & \textbf{Self} & \textbf{Total} & \textbf{Relationship} & \textbf{Self} & \textbf{Total} \\
\toprule
\multirow{2}*{Characteristic} &  1589  &  16431  &  \multirow{2}*{18020}  &  260  & 3886 &  \multirow{2}*{4146} & \multirow{2}*{22166} \\
& 7.2\% & 74.1\% & & 1.2\% & 17.5\% & & \\

\midrule
\multirow{2}*{Routine or Habit} &  13402  &  24248  &  \multirow{2}*{37650}  & 1429   & 8373  &  \multirow{2}*{9802} & \multirow{2}{*}{47452} \\
& 28.2\% & 51.1\% & & 3.0\% & 17.6\% & & \\

\midrule
\multirow{2}*{Goal or Plan}  & 3962	& 8956 & \multirow{2}*{12918} & 335 & 1562 & \multirow{2}*{1897} &\multirow{2}*{14815} \\
& 26.7\% & 60.5\% &	& 2.3\% & 10.5\% & & \\

\midrule
\multirow{2}*{Experience}  & 3089 & 11477 & \multirow{2}*{14566} & 427 & 2671 & \multirow{2}*{3098} & \multirow{2}*{17664}\\
& 17.5\% & 65.0\% & & 2.4\%	 & 15.1\% & & \\

\bottomrule
\end{tabular}
}
\caption{Fine-grained statistics of persona relations in \ourkg{}.}
\label{tab:fine_grained_stats}
\end{table*}

\begin{table}[t]
\centering
\resizebox{1.0\columnwidth}{!}{
\smallskip\begin{tabular}{l}
\toprule
\quad\;\textbf{\textit{Head}}: I am a programmer who become an expert\\
\cmidrule(lr){1-1}
\textbf{Relation}: Characteristic, Self, Distinctive \\
\quad\quad\textbf{\textit{Tail}}: tech savvy and highly knowledgeable in coding \\
\textbf{Relation}: Routine or Habit, Self, Distinctive \\
\quad\quad\textbf{\textit{Tail}}: write code and develop software \\
\textbf{Relation}: Goal or Plan, Self, Distinctive \\
\quad\quad\textbf{\textit{Tail}}: to create a new software application \\
\textbf{Relation}: Experience, Self, Distinctive \\
\quad\quad\textbf{\textit{Tail}}: earned a software engineering certification \\
\midrule
\quad\;\textbf{\textit{Head}}: I am a waiter\\
\cmidrule(lr){1-1}
\textbf{Relation}: Characteristic, Relationship, Distinctive \\
\quad\quad\textbf{\textit{Tail}}: skilled at customer service \\
\textbf{Relation}: Routine or Habit, Relationship, Distinctive \\
\quad\quad\textbf{\textit{Tail}}: get tips from customers \\
\midrule
\quad\;\textbf{\textit{Head}}: I am a great basketball player\\
\cmidrule(lr){1-1}
\textbf{Relation}: Goal or Plan, Relationship, Distinctive \\
\quad\quad\textbf{\textit{Tail}}: drafted by the NBA \\
\textbf{Relation}: Experience, Relationship, Distinctive \\
\quad\quad\textbf{\textit{Tail}}: played on the varsity basketball team in high school \\
\midrule
\quad\;\textbf{\textit{Head}}: I am a secure person \\
\cmidrule(lr){1-1}
\textbf{Relation}: Characteristic, Relationship, Generic \\
\quad\quad\textbf{\textit{Tail}}: important to family \\
\textbf{Relation}: Routine or Habit, Self, Generic \\
\quad\quad\textbf{\textit{Tail}}: receive compliment well \\
\midrule
\quad\;\textbf{\textit{Head}}: I am a runner who run track \\
\cmidrule(lr){1-1}
\textbf{Relation}: Goal or Plan, Self, Generic \\
\quad\quad\textbf{\textit{Tail}}: get better \\
\midrule
\quad\;\textbf{\textit{Head}}: I am a manager who work hard at my job \\
\cmidrule(lr){1-1}
\textbf{Relation}: Experience, Self, Generic \\
\quad\quad\textbf{\textit{Tail}}: get the job \\
\bottomrule
\end{tabular}
}
\caption{\ourkg{} examples of persona facts.}
\label{tab:example_persona_full}
\end{table}

\begin{table}[t]
\centering
\resizebox{1.0\columnwidth}{!}{
\smallskip\begin{tabular}{@{~}l@{~~~}l@{~}}
\toprule
\textbf{Relation}  & \textbf{Textual Description} \\
\toprule
Characteristic   & here is my character trait \\
Routine or Habit & here is what I regularly or consistently do \\
Goal or Plan     & here is what I will do or achieve in the future \\
Experience       & here is what I did in the past \\
Relationship     & related to other people or social groups \\
\bottomrule
\end{tabular}
}
\caption{Textual descriptions of relations in \ourkg{}.}
\label{tab:relation_peacok}
\end{table}

\section{Neural KG Analysis Details}
\label{sec:neural_analysis}

\paragraph{Fact Preprocessing}
We develop neural knowledge generator based on the \ourkg{} facts whose relations are labeled as \textit{Distinctive} in the third (distinctiveness) dimension.
We preprocess these distinctive \ourkg{} facts to facilitate knowledge generation.
In particular, we follow Table~\ref{tab:relation_peacok} to map each fact's relation into a textual description, and then concatenate it with the fact's \textit{head} and \textit{tail} personas.
If the relation is labeled as \textit{Relationship} in the second (interactivity) dimension, we also append its description in Table~\ref{tab:relation_peacok} to the fact's main-dimension label description, \ie{}, one of the other four descriptions in Table~\ref{tab:relation_peacok}.
For example, (\textit{I am a waiter}, \textit{Characteristic} and \textit{Relationship}, \textit{skilled at customer service}) is converted into \textit{I am a waiter, here is my character trait related to other people or social groups, skilled at customer service}.

\begin{figure*}[t]
\centering
\includegraphics[width=1.0\textwidth]{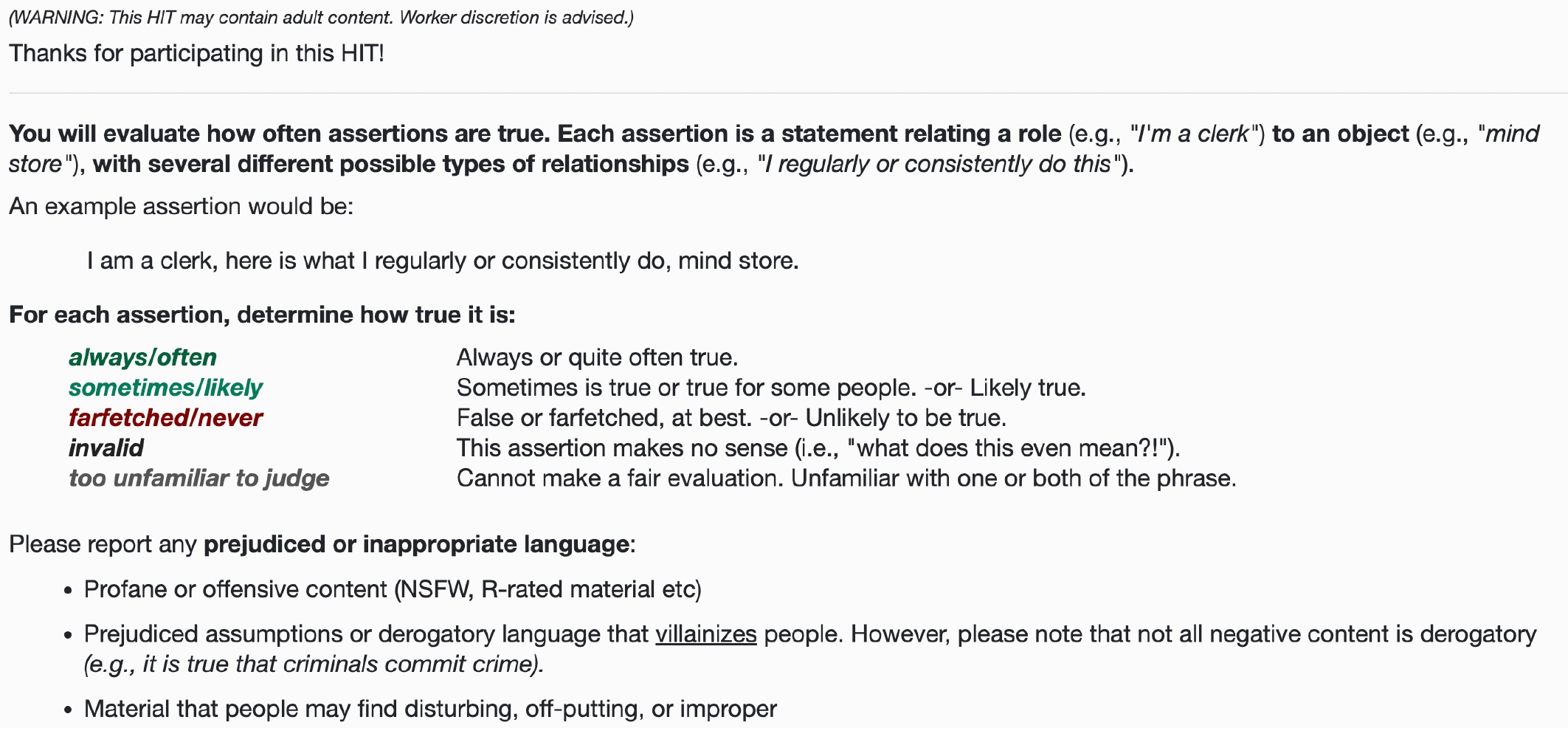}
\caption{Screenshot of our human evaluation instruction for neural KG analysis.}
\label{neural_crowd}
\end{figure*}

\paragraph{Evaluation Details}
We split our preprocessed facts into three sets, with size 64853, 8913 and 14112 for training, validation and testing, respectively.
Note that the three sets of facts do not have overlapped \textit{head} personas with each other.
We evaluate \textit{tail} persona generation on the 3030 unique \textit{head}-relation combinations in the testing set, with the 14112 gold \textit{tail} personas serving as references.
Several NLG metrics are adopted for the automatic evaluation, including cumulative 4-gram \textbf{BLEU} \citep{papineni2002bleu}, \textbf{ROUGE-L} \citep{lin2004rouge}, \textbf{METEOR} \citep{banerjee2005meteor} and \textbf{SkipThoughts} \citep{kiros2015skip}.
For human evaluation, we use the same group of workers qualified for \ourkg{} relation classification described in Appendix~\ref{sec:appendix_construction}.
Each fact with generated \textit{tail} is evaluated by one Amazon Mechanical Turk worker, following our instruction shown in Figure~\ref{neural_crowd}.
We pay each worker \$0.20 for evaluating every 5 facts, which keeps similar hourly wage as compared to \ourkg{} relation classification.

\paragraph{Model Training}
We use Kogito \citep{ismayilzada2022kogito} toolkit to train the \comet-BART knowledge generator, with the default hyperparameters suggested by the toolkit.
One NVIDIA TITAN X Pascal GPU is used to train the model for 7 epochs, which costs about 1 hour to get the highest ROUGE-L score on the validation set.
For the 5-shot GPT-3 generation, we prompt the davinci endpoint with default hyperparameters suggested by the OpenAI GPT-3 platform.

% DeBERTa training for binary analysis + results
We also train a DeBERTa \citep{he2020deberta} discriminator to re-rank the facts generated by \comet-BART and GPT-3.
For each training fact, we create one negative example by replacing its \textit{tail} persona with a randomly sampled one from another training fact, which have a different \textit{head} persona but same relation.
We train the DeBERTa model to discriminate true facts versus negative samples based on a binary classification loss, with hyperparameters suggested by the \comfact{} \citep{gao2022comfact} benchmark.
Four NVIDIA TITAN X Pascal GPUs are used to train the model for 6 epochs, which costs about 21 hours to get the highest F1 score on the validation set.
Finally, for both \comet-BART and GPT-3, we evaluate their top-1 of 5 generated facts re-ranked by our DeBERTa discriminator, with their default decoding methods, \ie{}, beam search for \comet-BART and nucleus sampling for GPT-3, with 1.0 top-p sampling rate and 0.9 temperature value.

\section{Persona Dialogue Agent Implementation Details}
\label{sec:appendix_downstream}
Our downstream dataset, ConvAI \personachat{}, contains 17878 and 1000  crowdsourced dialogues for training and validation, while 1015 testing dialogues are not public.
In each dialogue sample, two speakers are pre-given their own persona profiles, \ie{}, four or five sentences of self-introductions, to conduct conversations.
Based on the persona profiles, \psqbot{} uses a reinforcement learning \citep{sutton1999policy} approach to build mutual persona perception between speakers, which enhances the quality of personalized dialogue generation.

\paragraph{Persona Knowledge Linking}
We first link candidate facts from \ourkg{} via the pattern matching and embedding similarity heuristics introduced in \comfact{}, and then use a DeBERTa \citep{he2020deberta} entity linker trained on \comfact{} to select relevant facts from the candidates.
We use the DeBERTa entity linker (instead of fact linker) to check the relevance of each fact's \textit{head} and \textit{tail} personas independently, without considering their in-between relations.
This is because the DeBERTa fact linker from \comfact{} is trained on \atomicTT{} relations, which cannot well identify the new relation sets of \ourkg{}.
We link persona facts from \ourkg{} whose \textit{head} and \textit{tail} personas are both relevant to the extracted \personachat{} statement or sentence.
We also include an additional set of persona facts which only have relevant \textit{tail}, since the high-level \textit{head} personas are not always revealed in the dialogue.
Similar to the fact preprocessing described in Appendix~\ref{sec:neural_analysis}, we convert each linked persona fact into a natural language statement, by first following Table~\ref{tab:relation_peacok} to map each fact's relation into a textual description, and then concatenate it with the fact's \textit{head} and \textit{tail} personas.

\paragraph{Model Training}
We train our knowledge augmented models (\ie{}, \psqbot{} w/ \ourkg{} and \psqbot{} w/ \atomicTT{}) with the same hyperparameters and early stopping settings as the original \psqbot{} model.
Two NVIDIA TITAN X Pascal GPUs are used, which takes about 20 hours to get convergence (early stopped) on the validation set.

% \begin{figure*}[t]
% \centering
% \includegraphics[width=0.8\textwidth]{expert_qual.pdf}
% \caption{\textcolor{red}{Interface of our expert qualification test for human evaluation on downstream dialogue response generation.}}
% \label{expert_qual}
% \end{figure*}

\begin{table*}[t]
\centering
\resizebox{1.0\textwidth}{!}{
\smallskip\begin{tabular}{ll}
\toprule
\textbf{Evaluation Aspect}  & \textbf{Question} \\
\toprule
Fluency      & Which response is more fluent and understandable? \\
Consistency  & Which response is more consistent with the dialogue history? \\
Engagement   & Which response shows higher engagement, e.g., more attractive and interesting, more active involvement? \\
Persona Expression  & Which response shows richer personas of the interlocutor that are consistent with his or her persona profiles? \\
\bottomrule
\end{tabular}
}
\caption{Questions for human evaluation on downstream dialogue response generation, with regard to the four evaluation aspects.}
\label{tab:questions_downstream}
\end{table*}

\paragraph{Human Evaluation}
For each pairwise comparison, we show the experts two responses generated by different models, with the gold dialogue history and the interlocutor persona profiles. We ask the experts to compare the two responses with regard to our four evaluation aspects (\ie{}, fluency, consistency, engagement and persona expression). To guide the experts to better understand our evaluation criteria, we interpret each evaluation aspect as a specific question, as shown in Table~\ref{tab:questions_downstream}.

\paragraph{Downstream Dialogue Generation Examples}
Table~\ref{tab:downstream_examples} presents an example of our downstream dialogue generation results, where we show the response generated by each model along with the dialogue history and the persona profile associated with the speaker of the response.
The linked \ourkg{} knowledge (\ie{}, fact) that involved in the response generation is also presented.
We find that the involved \ourkg{} fact help identify a potential role of the speaker, \ie{}, \textit{breeder} inferred from \textit{milking cows} and \textit{farmland}, and also explain the speaker's persona of having \textit{a pet canine}, \ie{}, \textit{dog}.
Therefore, compared to other baseline models, \psqbot{} w/ \ourkg{} generates a more consistent and engaging response, which is well associated with the counterpart's last utterance in the dialogue history, and also simultaneously conveys a related persona of the speaker.

\begin{table*}[t]
\centering
\resizebox{1.0\textwidth}{!}{
\smallskip\begin{tabular}{crl}
\toprule
\multicolumn{2}{c}{\multirow{5}*{Dialogue History}}   & Hi there, how are you tonight? \\
                                  & & I'm tired from milking the cows. \\
                                  & & Well, I know I am strong for a 100, but I don't think I can do that. \\
                                  & & 100?! That's older than my dad! Would you make your son milk cows? \\
                                  & & Yeah because I sure can't do it. \\
\midrule
\multicolumn{2}{c}{\multirow{4}*{Persona Profile}}   & The Black Hawks are my favorite team! \\
                                  & & I reside in the country surrounded by farmland. \\
                                  & & I like to meditate with art. \\
                                  & & I have a pet canine that looks like a police canine. \\
\midrule
\multirow{4}*{Responses}          & \textbf{\psqbot{}}: & I' m sorry. I am a Buddhist, but I like to meditate. \\
                                  & \textbf{\psqbot{} w/ \atomicTT{}}: & Well, I'm a grown man, and I do not want to be a cop.  \\
                                  & \textbf{\psqbot{} w/ \ourkg{}}: & You can do it! I have a dog. He can help you. \\
                                  & \textbf{Human}: & I would rather be coloring in my books! Do you like sports? \\
\midrule
\multicolumn{2}{c}{Involved \ourkg{} Knowledge} & I am a breeder, Routine or Habit, breed dog \\
\bottomrule
\end{tabular}
}
\caption{An example of downstream dialogue response generation.}
\label{tab:downstream_examples}
\end{table*}

\end{document}